%% file: ORCA_v2.tex
\documentclass{article}

\PassOptionsToPackage{numbers, compress}{natbib}

\usepackage[preprint, main]{neurips_2026}
\usepackage[utf8]{inputenc} 
\usepackage[T1]{fontenc}    
\usepackage{hyperref}       
\usepackage{url}            
\usepackage{booktabs}       
\usepackage{amsfonts}       
\usepackage{nicefrac}       
\usepackage{microtype}      
\usepackage{xcolor}         

\usepackage{amsmath, amssymb, amsthm}
\usepackage{bm}
\newtheorem{theorem}{Theorem}

\usepackage{algorithm}
\usepackage{algpseudocode}

\usepackage{graphicx}
\usepackage{array}
\usepackage{multirow}
\usepackage{subfig}        
\usepackage{stfloats}

\usepackage{textcomp}
\usepackage{verbatim}

\begin{document}

\title{ORCA: Orthogonal Residual Consensus Alignment for Multi-View Clustering}

\author{%
	Xiaojian Ding,  Xian Li, Xiaoying Zhu, Lin Zhao, Kaixiang Wang\thanks{Corresponding author}\\
School of Computer and Artificial Intelligence,\\
 Nanjing University of Finance and Economics, 
Nanjing, China \\
xjding@nufe.edu.cn, 1120241388@stu.nufe.edu.cn,\\ 1120241394@stu.nufe.edu.cn, 1120241386@stu.nufe.edu.cn,\\ kxwang@nufe.edu.cn
}

\maketitle

\begin{abstract}
Deep multi-view clustering often suffers from redundant information accumulation during fusion and indiscriminate alignment across representation levels. We propose Orthogonal Residual Consensus Alignment (ORCA) to address these challenges. Its semantic incremental recurrent fusion module extracts cross-view orthogonal residuals as non-redundant increments and adaptively weights recurrent states through divergence-guided gating. Its consensus-calibrated hierarchical contrastive alignment module uses the global consensus to weight contrastive alignment of view-level, pairwise, and recurrent representations. Together, these modules balance view-specific structure with cross-view consistency. Experiments on five benchmarks against fifteen competing methods demonstrate ORCA’s effectiveness, including a 5.25\% accuracy improvement over the strongest baseline on 100leaves. Ablation studies validate both modules. Code is available at \url{https://github.com/XiaojianDing/2026-NeurIPS-ORCA}.

\end{abstract}

\section{Introduction}

With the ubiquity of heterogeneous data, Multi-View Clustering (MVC) has emerged as a fundamental research topic in unsupervised learning by integrating consistency and complementarity across diverse views \cite{li2026deep}. Although deep learning has significantly enhanced the ability to model nonlinear correlations, constructing compact and discriminative consensus representations remains a core challenge due to inherent issues such as information redundancy and view bias. Consequently, exploring robust multi-view fusion and alignment strategies is essential for achieving precise data partitioning and profound semantic understanding \cite{xu2026bridging}.

Recent research in deep multi-view clustering (MVC) has shifted from simple multi-view feature aggregation toward more sophisticated semantic consistency modeling and alignment. Recent advancements highlight the efficacy of multi-level contrastive learning, exemplified by frameworks such as 3MC \cite{chen2025multi}, MCMC \cite{bian2025multilevel}, and DMvCGSA \cite{fei2025deep}. By establishing a hierarchical representation structure that integrates instance-, cluster-, and prototype-level information, these models substantially improve the compactness and discriminative power of learned features. Simultaneously, there is a growing interest in structured graph representation learning, with methods like LFCGL \cite{wang2026locality}, SemGProp \cite{qiu2026self}, and MISCC \cite{li2026deep} utilizing graph convolutional networks or self-supervised graph propagation to model topological relationships and improve robustness against noise. Furthermore, researchers are increasingly exploring disentangled representation learning and redundancy elimination; for instance, STMVC \cite{ren2026enhancing} removes cross-view redundancy through an information-theoretic lens, while SCDFL \cite{zhang2024separable} and DMCAR-MVC \cite{xiao2026dmcar} explicitly decouple shared consensus semantics from view-specific information to mitigate representation collapse.

Despite recent progress in deep multi-view clustering, existing methods still suffer from two key limitations. First, many fusion-based approaches aggregate heterogeneous view representations in a redundancy-blind manner, where shared and complementary information are not explicitly distinguished. As a result, repetitive cross-view components may be repeatedly propagated during feature fusion, while truly discriminative view-specific semantics are diluted. This issue becomes more pronounced in attention-based fusion, since iterative aggregation tends to contract the latent representation space and may lead to over-smoothed features or even representation collapse, weakening the separability of clustering boundaries. Although recent methods introduce contrastive objectives or attention mechanisms to enhance cross-view consistency, they usually rely on learned weights to implicitly suppress redundancy, without explicitly modeling the non-redundant semantic residuals between views. 
Second, existing alignment strategies often treat multi-view semantics at a single level or align different representations with uniform constraints, ignoring the hierarchical nature of cross-view information. In practice, useful clustering cues may appear at multiple semantic levels, including individual view structures, local pairwise relations, and intermediate states produced during recurrent interaction. However, directly enforcing these representations to align with the final consensus may introduce noisy or weakly relevant semantics into the optimization process. Although multi-scale or contrastive learning methods have been explored, they generally lack a consensus-anchored calibration mechanism to evaluate which semantic levels are more consistent with the global clustering structure. Consequently, the learned representation may fail to simultaneously preserve local structural details and maintain globally discriminative consensus semantics.

To address the above challenges, we propose Orthogonal Residual Consensus Alignment (ORCA), a unified framework for robust multi-view clustering. ORCA is built upon two complementary modules: Semantic Incremental Recurrent Fusion (SIRF) and Consensus-Calibrated Hierarchical Contrastive Alignment (CCHA). Instead of directly aggregating heterogeneous view representations, SIRF first identifies non-redundant semantic increments from cross-view orthogonal residuals, enabling each view to absorb complementary information while suppressing repetitive components. It further formulates multi-view integration as a recurrent semantic interaction process, where the evolution of fused representations is regulated by information gain to alleviate redundant assimilation and representation collapse. Meanwhile, CCHA constructs a hierarchical semantic pool from view-level, local pairwise, and recurrent interaction representations, and uses the learned global consensus as an anchor to adaptively calibrate their contributions through attention-weighted contrastive alignment. In this way, ORCA jointly realizes redundancy-aware fusion, consensus formation, and hierarchical semantic calibration within a single optimization framework.

The main contributions of this work are:
\begin{itemize}
    \item We propose a geometric projection-based strategy to explicitly separate redundant and complementary semantics, adaptively enriching core views with non-redundant increments to reduce information accumulation.
    \item We design a calibration mechanism that uses global consensus as an anchor to adaptively weight and align multi-level semantics, ensuring consistent and discriminative clustering representations.
    \item Evaluations across diverse modalities demonstrate that ORCA outperforms fifteen state-of-the-art baselines, with ablation and visualization results confirming its robust discriminative capability.
\end{itemize}

\section{Related Work}
\paragraph{Feature-Alignment-Based Contrastive MVC.} 
Early methods utilize autoencoders to extract view-specific features and align them in latent spaces via contrastive losses. Examples include DealMVC \cite{yang2023dealmvc} and SCMC \cite{fu2024subspace}, which enforce local or subspace contrastive constraints. MCoCo \cite{zhou2024mcoco} and DFMVC \cite{zhao2024dfmvc} extend this to the semantic label space for category-level alignment, while CDARL \cite{xi2025contrastive} employs dual adversarial and contrastive learning to minimize distribution discrepancies. Moreover, methods like TDLSR \cite{li2026theory} develop theory-driven label-specific representations to robustly align features even under incomplete data scenarios. To improve efficiency, DWCL \cite{yuan2026dwcl} introduces a "Best-Other" mechanism to select reliable views. However, these methods often apply uniform contrastive intensity to all views, failing to distinguish between shared and view-specific information, which results in redundant information propagation.

\paragraph{Multi-Level and Multi-Granularity Contrastive Learning.} 
Recent studies explore hierarchical structures to capture multi-granularity semantics. 3MC \cite{chen2025multi} performs layer-wise feature and label alignment, while MCMC \cite{bian2025multilevel} proposes a triple-contrastive framework across instance, cluster, and prototype levels. DMCAR-MVC \cite{xiao2026dmcar} utilizes a decoupled Mixture-of-Experts (MoE) architecture with orthogonal constraints to separate shared and specific representations. Additionally, PSGVC \cite{li2026pseudo} leverages pseudo-label similarity graphs to guide global and local alignment. Despite their success, these methods lack a global consensus anchor to calibrate the contributions of different semantic levels, often leading to the indiscriminate alignment of noisy or weakly relevant semantics.

\paragraph{Fusion-Driven Consensus Learning and Redundancy Suppression.} 
Effective fusion is critical for DMVC. While methods like GCFAgg \cite{yan2023gcfagg} and MISCC \cite{li2026deep} focus on aggregating complementary info through cross-view correlations, others emphasize redundancy suppression. GMAE \cite{zou2026learning} decouples features into view-specific and view-common embeddings, whereas STMVC \cite{ren2026enhancing} uses an information-theoretic approach to eliminate redundant components. Parallelly, approaches like AFIP \cite{li2026correcting} theoretically prove that calibrating attention to correct visual distraction is necessary to reduce noisy implications during feature integration. Furthermore, ILASC \cite{yang2026deep} adaptively fuses views using Transformer Pooling and aligns the overall clustering structure. Nevertheless, most fusion strategies lack explicit modeling of non-redundant semantic residuals, which can dilute complementary information.

\paragraph{Pseudo-Label Guided and Structure-Aware Contrastive Learning.} 
To mitigate unreliable positive/negative pairs and noisy annotations, researchers have introduced external guidance. For instance, semi-supervised paradigms have been enhanced by unbiased losses in DIMvSML \cite{li2024deep}, view-specific transformers guided by pseudo-labels in SMVTEP \cite{li2025semi}, and theory-inspired noise-resistant frameworks like DMMIvNL \cite{li2025theory}. Additionally, PBDMSC \cite{yu2025pseudo} uses cluster assignments to select reliable samples, and STCMC-UR \cite{hu2025self} filters false negatives via uncertainty estimation. SemGProp \cite{qiu2026self} propagates high-confidence pseudo-label structures to enhance consistency, while CGM-CUC \cite{dong2026cross} addresses unaligned scenarios through graph matching and iterative alignment. While these approaches improve signal reliability, they typically focus on single-level alignment and lack a systematic mechanism to calibrate hierarchical semantics against the global clustering goal.

\section{Method}

\begin{figure*}[!t]
    \centering
    \includegraphics[width=0.95\textwidth]{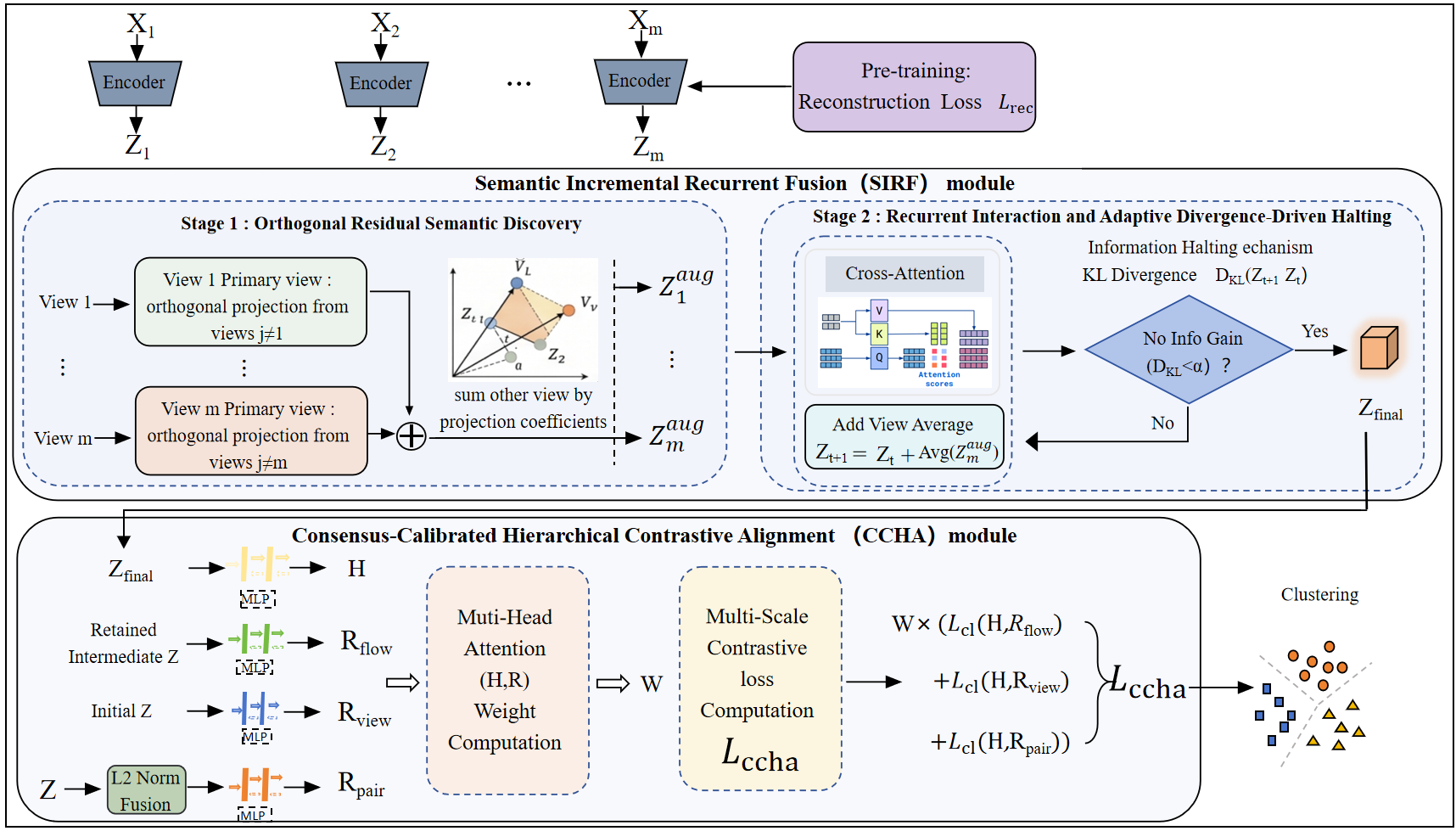}
    \caption{Overview of the proposed ORCA framework, consisting of three main components: (i) View-specific Autoencoders for latent feature extraction; (ii) The SIRF module, which extracts complementary residuals to distill a non-redundant global consensus $\mathbf{H}$; and (iii) The CCHA module, which adaptively aligns multi-scale representations (view-level, local, and recurrent) against the global consensus anchor.}
    \label{fig:architecture}
\end{figure*}

In this section, we detail the proposed ORCA framework, as shown in Figure ~\ref{fig:architecture}. Our architecture relies on two coordinated core modules. First, the SIRF module explicitly separates shared redundancy from complementary residual semantics via geometric projection. By autonomously halting the semantic interaction, SIRF maximizes information gain without risking structural collapse. Second, the CCHA module is introduced to preserve multi-scale semantics. It constructs a hierarchical pool (spanning view-level, local pairwise, and recurrent representations) and adaptively aligns these features against a global consensus anchor via attention-guided contrastive learning, ultimately yielding consistent and robust clustering representations.

\subsection{Semantic Incremental Recurrent Fusion}

In multi-view clustering, existing attention-based fusion paradigms inherently act as low-pass filters, shrinking the latent space and causing severe dimensionality collapse. To overcome this, we propose the SIRF module, which formulates multi-view aggregation as a dynamical system. SIRF operates through two core stages: orthogonal residual semantic discovery and adaptive divergence-driven halting. By explicitly extracting topology-aware semantic increments and autonomously halting the recurrent interaction, SIRF prevents spectral degeneration and maximizes information gain, ultimately yielding a highly discriminative global representation.
\subsubsection{Orthogonal Residual Semantic Discovery}
Before mapping multiple views into a shared semantic space, we must recognize that different views often share significant overlapping information. Direct aggregation introduces massive feature redundancy, which overwhelms the view-specific discriminative structures. To explicitly isolate the truly complementary semantic information contributed by supplementary views, we propose an orthogonal residual semantic discovery strategy.
From a geometric perspective, if we treat the features as vectors in a high-dimensional manifold, the component of a marginal view $\mathbf{Z}_{\mathrm{other}}$ that is parallel to the core view $\mathbf{Z}_{\mathrm{core}}$ represents redundant, already-known information. Conversely, the strictly orthogonal residual component constitutes the non-redundant semantic supplement. 

Given $V$ distinct initial views, we systematically treat each view $v$ as the core view $\mathbf{Z}_{\mathrm{core}}^{(v)}$, and the remaining views $u \in \{1, \dots, V\} \setminus \{v\}$ as the marginal views $\mathbf{Z}_{\mathrm{other}}^{(u)}$. We first project the marginal view onto the direction of the core view. This projection $\mathbf{Z}_{\mathrm{proj}}^{(u \to v)}$, which encapsulates the structural consensus, is calculated as:
\begin{equation}
\mathbf{Z}_{\mathrm{proj}}^{(u \to v)} = \left( \frac{\langle \mathbf{Z}_{\mathrm{other}}^{(u)}, \mathbf{Z}_{\mathrm{core}}^{(v)} \rangle}{\|\mathbf{Z}_{\mathrm{core}}^{(v)}\|_2^2} \right) \mathbf{Z}_{\mathrm{core}}^{(v)}.
\end{equation}

To extract the pure novel information independent of the core view, we compute the orthogonal residual component by subtracting the parallel projection from the marginal view:
\begin{equation}
\mathbf{Z}_{\bot}^{(u \to v)} = \mathbf{Z}_{\mathrm{other}}^{(u)} - \mathbf{Z}_{\mathrm{proj}}^{(u \to v)}.
\end{equation}

By measuring the relative strength of orthogonal residuals, the proposed method adaptively enriches each core view with non-redundant semantic increments, thereby reducing repetitive information accumulation in multi-view fusion. We define a scalar residual weight $g^{(u \to v)}$ as the relative magnitude of the orthogonal component:
\begin{equation}
g^{(u \to v)} = \frac{\|\mathbf{Z}_{\bot}^{(u \to v)}\|_2}{\|\mathbf{Z}_{\mathrm{other}}^{(u)}\|_2}.
\end{equation}
A larger $g^{(u \to v)}$ explicitly indicates that the marginal view introduces a higher proportion of complementary residual semantics. Finally, we inject this weighted orthogonal residual into the core view to construct the enhanced, information-maximized initial representation $\hat{\mathbf{Z}}^{(v)}$:
\begin{equation}
\hat{\mathbf{Z}}^{(v)} = \mathbf{Z}_{\mathrm{core}}^{(v)} + \sum_{u \neq v} g^{(u \to v)} \cdot \mathbf{Z}_{\bot}^{(u \to v)}.
\end{equation}
This rigorous geometric stripping completely abandons empirical guessing, ensuring that the enhanced feature set $\mathbf{Z}^{(0)} = \{\hat{\mathbf{Z}}^{(1)}, \dots, \hat{\mathbf{Z}}^{(V)}\}$ spans a maximally informative and non-redundant subspace based on complementary semantics rather than repetitive information.

\subsubsection{Recurrent Interaction and Adaptive Divergence-Driven Halting}
Using the information-enhanced feature embeddings $\mathbf{Z}^{(0)}$, we construct a cross-view interaction space to model global contextual dependencies. To prevent the semantic flow from drifting during deep interactions and to anchor the representations to a stable structural baseline, we compute the global mean of the initial raw views as $\bar{\mathbf{Z}} = \frac{1}{V} \sum_{v=1}^V {\mathbf{Z}}^{(v)}_{\mathrm{raw}}$. At the $t$-th iteration of our dynamic interaction system, the intermediate cross-view semantic representation $\mathbf{Z}^{(t)}$ is iteratively updated via a self-attention mechanism combined with this explicit structural residual:
\begin{equation}
\mathbf{Z}^{(t)} = \mathrm{Softmax}\left( \frac{\mathbf{Z}^{(t-1)} \mathbf{W}_Q (\mathbf{Z}^{(t-1)} \mathbf{W}_K)^{\top}}{\sqrt{d_k}} \right) \mathbf{Z}^{(t-1)} \mathbf{W}_V + \bar{\mathbf{Z}}.
\end{equation}

While this attention operator effectively establishes long-range correlations across different views, its softmax-induced convex combination fundamentally induces metric distortion. Over successive recurrent aggregations, the spectral radius of the attention Jacobian matrix is strictly bounded below 1. Even with the injection of the global residual $\bar{\mathbf{Z}}$, this contractive mapping ultimately behaves as a low-pass filter over deep iterations, leading to catastrophic dimensionality collapse where clustering margins are obliterated. 

To counteract this topological degeneration without disrupting the parallel efficiency of batch training, we shift from heuristic structural constraints to an information-theoretic soft-gating mechanism. We dynamically monitor the instantaneous Information Gain during the semantic interaction process using the Kullback-Leibler (KL) divergence. Let $\mathcal{P}(\mathbf{Z}^{(t)})$ denote the feature distribution of the system at iteration $t$. The instantaneous information gain $\Delta \mathcal{I}^{(t)}$ is quantified as:
\begin{equation}
\Delta \mathcal{I}^{(t)} = \mathcal{D}_{KL}\Big( \mathcal{P}(\mathbf{Z}^{(t)}) \parallel \mathcal{P}(\mathbf{Z}^{(t-1)}) \Big).
\end{equation}

Drawing inspiration from adaptive computation time frameworks, we map this information gain to an adaptive retention weight $\alpha_t \in (0, 1)$ through a learnable gating function:
\begin{equation}
\alpha_t = \sigma\left( \mathbf{W}_{\alpha} \Delta \mathcal{I}^{(t)} + \mathbf{b}_{\alpha} \right),
\end{equation}
where $\sigma(\cdot)$ is the Sigmoid activation, and $\mathbf{W}_{\alpha}$, $\mathbf{b}_{\alpha}$ are trainable parameters. This mechanism acts as a soft-halting gate: when the system actively learns new consensus semantics ($\Delta \mathcal{I}^{(t)}$ is large), the interaction continues; as the information exchange saturates ($\Delta \mathcal{I}^{(t)} \to 0$), $\alpha_t$ smoothly decays, effectively terminating the repetitive assimilation of saturated states.

The model synthesizes the optimal consensus representation by accumulating the dynamically weighted states across all $T_{\mathrm{max}}$ potential iterations:
\begin{equation}
\mathbf{Z}_{\mathrm{final}} = \sum_{t=1}^{T_{\mathrm{max}}} \alpha_t \mathbf{Z}^{(t)}.
\end{equation}

Ultimately, to project this information-maximized, dynamically halted consensus representation into a highly discriminative latent space for downstream tasks, we pass $\mathbf{Z}_{\mathrm{final}}$ through a multi-layer perceptron. This yields the final global semantic representation $\mathbf{H}_{\mathrm{final}}$:
\begin{equation} \label{eq:global_representation}
\mathbf{H}_{\mathrm{final}} = \mathrm{MLP}(\mathbf{Z}_{\mathrm{final}}; \psi),
\end{equation}
where $\psi$ represents the learnable parameters. By continuously capturing cross-view consensus and autonomously halting the flow before the onset of structural collapse, $\mathbf{H}_{\mathrm{final}}$rigorously preserves intrinsic information volume while establishing clear boundaries for clustering.

\subsection{Consensus-Calibrated Hierarchical Contrastive Alignment}

While the SIRF module effectively distills a global consensus $\mathbf{H}$, relying solely on the final layer output may neglect rich hierarchical semantics embedded in the fusion trajectory. To avoid indiscriminate constraints in multi-scale semantic alignment, we design a consensus-anchored hierarchical semantic calibration strategy. By constructing a hierarchical semantic pool and using the global consensus representation as the semantic anchor, CCHA adaptively weights different semantic levels through multi-head attention and performs weighted contrastive alignment, ensuring that $\mathbf{H}$ encapsulates both robust local details and consistent global patterns.

\subsubsection{Hierarchical Semantic Pool Construction}
Before applying contrastive objectives, we systematically construct a hierarchical semantic pool denoted as $\mathcal{Z}_{\text{pool}}$. In the context of our CCHA module, ``semantics'' refers to the manifold topological structures, latent clustering cues, and discriminative features captured across different representation stages and receptive fields. We preserve these multi-scale semantics because single-level representations often struggle to balance local feature integrity with global clustering consistency. By capturing the data topology across multiple receptive fields, this hierarchy comprehensively characterizes the evolutionary trajectory of information from isolation to assimilation across three distinct tiers:

\textbf{View-level scale ($\mathcal{S}_{\text{view}}$):} We directly preserve the raw structural information from each individual view's latent embedding, yielding $\mathcal{S}_{\text{view}} = \{\mathbf{z}_v\}_{v=1}^V$. This foundational tier retains the initial, independent structures of individual views to fundamentally prevent the loss of discriminative, view-specific information.

\textbf{Local pairwise scale ($\mathcal{S}_{\text{pair}}$):} To explicitly capture cross-view complementarity and bridge local view discrepancies, every unordered pair $(i, j)$ of the initial embeddings is fused using $\ell_2$-norm based adaptive weights:
\begin{equation}
\omega_i = \frac{\|\mathbf{z}_i\|_2}{\|\mathbf{z}_i\|_2 + \|\mathbf{z}_j\|_2}, \qquad \omega_j = 1 - \omega_i.
\end{equation}
This yields a set of locally fused representations $\mathcal{S}_{\text{pair}} = \{ \omega_i \mathbf{z}_i + \omega_j \mathbf{z}_j \mid 1 \leq i < j \leq V \}$, which explicitly records local pairwise interactions and complementary semantic correlations.

\textbf{Recurrent interaction scale ($\mathcal{S}_{\text{flow}}$):} During the previous divergence-guided recurrent fusion process, the cross-attention mechanism generates a trajectory of intermediate states $\mathbf{Z}^{(t)}$ driven by instantaneous information gain. To capture this progressive assimilation of multi-view semantics and record the dynamic formation of a global consensus while preventing semantic drift, we anchor these intermediate states to a stable global baseline. Specifically, we compute the global mean anchor of the initial raw views, $\bar{\mathbf{Z}} = \frac{1}{V}\sum_{v=1}^V \mathbf{z}_v$, and inject it as a structural residual into each intermediate state. This yields the recurrent interaction set:
\begin{equation}
\mathcal{S}_{\text{flow}} = \left\{ \mathbf{Z}^{(t)} + \bar{\mathbf{Z}} \mid t = 1, \dots, T_{\max} \right\}.
\end{equation}

We aggregate all the above multi-scale features into a unified candidate pool $\mathcal{Z}_{\text{pool}} = \mathcal{S}_{\text{view}} \cup \mathcal{S}_{\text{pair}} \cup \mathcal{S}_{\text{flow}}$. To align them into a shared semantic subspace for contrastive learning, every candidate $\mathbf{z}_m \in \mathcal{Z}_{\text{pool}}$ is projected through a shared multi-layer perceptron:
\begin{equation}
\mathbf{R}_m = \mathrm{MLP}(\mathbf{z}_m; \Theta), \qquad m = 1, \dots, M,
\end{equation}
where $M$ is the total number of candidate representations across all three scales (i.e., $M = V + \frac{V(V-1)}{2} + T_{\max}$).

Constructing a multi-scale semantic pool offers clear advantages over traditional contrastive frameworks. Single-scale alignment may introduce noise and dilute view-specific clustering cues. In contrast, our hierarchical pool preserves semantic information across isolated views, local synergies, and global consensus, providing diverse anchors for attention-based selection. This enables adaptive assessment of each representation's relevance to the global clustering objective, mitigating topological degradation while preserving local details and enhancing global cluster discrimination.

\subsubsection{Consensus-Anchored Calibration Weighting}
Not all multi-scale representations contribute equally to the final consensus. To adaptively regulate the influence of each representation $\mathbf{R}_m$ during contrastive alignment, we utilize the global consensus representation as the semantic anchor for our multi-head attention mechanism. Unlike standard self-attention, we use the global consensus $\mathbf{H}$ as the Query to calculate the correlation between different level representations and the clustering consensus.
For an attention head $h \in \{1, \dots, H_{att}\}$, we compute the linear projections:
\begin{equation}
\mathbf{Q}^{(h)} = \mathbf{H} W_Q^{(h)}, \quad \mathbf{K}_m^{(h)} = \mathbf{R}_m W_K^{(h)},
\end{equation}
where $W_Q^{(h)}, W_K^{(h)} \in \mathbb{R}^{d \times d_k}$ are learnable parameters. The attention alignment score between the global consensus anchor $\mathbf{H}$ and the $m$-th scale representation $\mathbf{R}_m$ is computed as:
\begin{equation}
\alpha_m^{(h)} = \frac{\exp\left( \langle \mathbf{Q}^{(h)}, \mathbf{K}_m^{(h)} \rangle / \sqrt{d_k} \right)}{\sum_{t=1}^M \exp\left( \langle \mathbf{Q}^{(h)}, \mathbf{K}_t^{(h)} \rangle / \sqrt{d_k} \right)}.
\end{equation}
To obtain a comprehensive scale-level importance weight, we average the attention intensity across all $H_{att}$ heads: $r_m = \frac{1}{H_{att}} \sum_{h=1}^{H_{att}} \alpha_m^{(h)}$. The final normalized contrastive weight for each scale is derived via softmax:
\begin{equation}
w_m = \frac{\exp(r_m)}{\sum_{t=1}^M \exp(r_t)}, \qquad m = 1, \dots, M.
\label{eq:ccha_weight}
\end{equation}
A higher $w_m$ indicates that the specific representation $\mathbf{R}_m$ is closer to the global clustering structure, thereby warranting a stronger constraint during alignment.

\subsubsection{Unified Objective Optimization}
To jointly optimize the global consensus and the multi-scale representations, we decompose the consensus-calibrated hierarchical contrastive alignment loss ($\mathcal{L}_{\mathrm{ccha}}$) into three distinct scale-specific objectives. Let $\mathcal{I}_{\text{view}}$, $\mathcal{I}_{\text{pair}}$, and $\mathcal{I}_{\text{flow}}$ denote the index sets for the candidate representations belonging to the view-level ($\mathcal{S}_{\text{view}}$), local pairwise ($\mathcal{S}_{\text{pair}}$), and recurrent interaction ($\mathcal{S}_{\text{flow}}$) scales, respectively. 

Utilizing the dynamically calibrated weights $w_m$ derived from the global anchor, representations from each specific scale are forced to align with the global consensus $\mathbf{H}$ using the weighted InfoNCE objective. The three scale-specific alignment losses are formulated as follows:

\begin{equation}
\mathcal{L}_{\mathrm{view}} = - \sum_{m \in \mathcal{I}_{\text{view}}} \sum_{i} w_m \log 
\frac{\exp\left( \mathrm{sim}(\mathbf{R}_m^i, \mathbf{H}^i) / \tau \right)}
{\sum_k \exp\left( \mathrm{sim}(\mathbf{R}_m^i, \mathbf{H}_k^{\text{neg}}) / \tau \right)},
\end{equation}

\begin{equation}
\mathcal{L}_{\mathrm{pair}} = - \sum_{m \in \mathcal{I}_{\text{pair}}} \sum_{i} w_m \log 
\frac{\exp\left( \mathrm{sim}(\mathbf{R}_m^i, \mathbf{H}^i) / \tau \right)}
{\sum_k \exp\left( \mathrm{sim}(\mathbf{R}_m^i, \mathbf{H}_k^{\text{neg}}) / \tau \right)},
\end{equation}

\begin{equation}
\mathcal{L}_{\mathrm{flow}} = - \sum_{m \in \mathcal{I}_{\text{flow}}} \sum_{i} w_m \log 
\frac{\exp\left( \mathrm{sim}(\mathbf{R}_m^i, \mathbf{H}^i) / \tau \right)}
{\sum_k \exp\left( \mathrm{sim}(\mathbf{R}_m^i, \mathbf{H}_k^{\text{neg}}) / \tau \right)},
\end{equation}
where $i$ denotes the sample index within a batch, $\mathbf{H}_k^{\text{neg}}$ represents negative consensus samples, and $\tau$ is the temperature scaling parameter.

The comprehensive consensus-calibrated hierarchical contrastive alignment loss is then constructed by integrating these three components:
\begin{equation}
\mathcal{L}_{\mathrm{ccha}} = \alpha \mathcal{L}_{\mathrm{view}} + \beta \mathcal{L}_{\mathrm{pair}} + \gamma \mathcal{L}_{\mathrm{flow}},
\label{eq:ccha_loss}
\end{equation}
where $\alpha$, $\beta$, and $\gamma$ are trade-off hyperparameters designed to balance the macro level contributions of the view-level, local pairwise, and recurrent interaction scales, while $w_m$ regulates the micro level importance of individual representations.

To simultaneously ensure that the view-specific encoders $E_v(\cdot)$ and decoders $D_v(\cdot)$ preserve essential intra-view semantic structures without distortion, a standard reconstruction regularization term is included:
\begin{equation}
\mathcal{L}_{\mathrm{rec}} = \sum_{v=1}^V \left\| \mathbf{X}_v - D_v(E_v(\mathbf{X}_v)) \right\|_F^2.
\end{equation}

The overall training objective of the network integrates the reconstruction constraints with the dynamically weighted hierarchical alignment:
\begin{equation}
\mathcal{L}_{\mathrm{total}} = \mathcal{L}_{\mathrm{rec}} + \lambda \mathcal{L}_{\mathrm{ccha}},
\label{eq:new_total_loss}
\end{equation}
where $\lambda$ is a balancing hyperparameter. 

After the network converges, clustering is performed directly on the optimal global consensus $\mathbf{H} \in \mathbb{R}^{N \times d}$ using $k$-means:
\begin{equation}
\min_{\mathbf{U}, \mathbf{C}} \left\| \mathbf{H} - \mathbf{U} \mathbf{C} \right\|_F^2, \quad \text{s.t.} \quad \mathbf{U} \mathbf{1} = \mathbf{1}, \quad \mathbf{U} \geq \mathbf{0}.
\end{equation}
\section{Experiments}
\subsection{Experimental Setup}

\textbf{Datasets:} We evaluate our method on five diverse multi-view benchmarks. Fashion~\cite{xu2021multi}, Synthetic3D~\cite{kumar2011co}, and Cifar10~\cite{chauhan2018convolutional} serve as representative vision datasets featuring distinct visual descriptors. MNIST-USPS~\cite{savaram2024comparative} provides complementary views for cross-domain digit clustering, while 100leaves~\cite{li2021consensus} encompasses diverse botanical features (shape, texture, and margin) for plant species clustering.

\textbf{Implementation Details:} To ensure fair comparisons and reproducibility, we use a unified model configuration for all datasets. Input data are vectorized before training. For feature extraction, each view employs an identical fully connected (FC) autoencoder with an encoder of four layers: Input -- FC500 -- FC500 -- FC2000 -- FC64, and a mirrored decoder. Following the extraction of the 64-dimensional latent embeddings, we construct the hierarchical semantic pool encompassing the view-level, local pairwise, and recurrent interaction scales. To align these multi-scale features into a shared semantic subspace for contrastive learning, every candidate representation within the pool is fed into a shared lightweight MLP (FC64 -- FC20) to obtain the projected representation $\mathbf{R}_m$. ReLU activations follow each hidden layer except the output layers. The model is trained using Adam optimizer with a fixed learning rate of 0.0003 and no weight decay or learning rate scheduler. Experiments run on a workstation with Intel i7-12700K CPU, 32 GB RAM, and NVIDIA RTX 4070 GPU using PyTorch 1.10.2.

\textbf{Comparison Methods:} To demonstrate the superiority of our proposed method, we evaluate it against fifteen state-of-the-art deep multi-view clustering baselines. 
Specifically, CSOT~\cite{zhang2024learning} employs optimal transport for cross-view semantic alignment, while DealMVC~\cite{yang2023dealmvc} captures discriminative features via pseudo-label graph learning. 
Several methods utilize contrastive learning frameworks: SCM~\cite{luo2024simple} integrates it with graph alignment, CVCL~\cite{chen2023deep} optimizes it for large-scale scenarios, SCMVC~\cite{wu2024self} combines it with spectral clustering, SEM~\cite{xu2023self} enhances it with semantic information, MCMC~\cite{bian2025multilevel} captures multilevel representational structures and enhances instance-level compactness via a dual self-supervised learning framework, and the recent HCLAA~\cite{ding2026hclaa} achieves fine-grained semantic decoupling through hierarchical contrastive learning and a dual-path attention mechanism. 
In terms of feature representation, DDMVC~\cite{xu2025deep} focuses on discriminative latent learning, MFLVC~\cite{xu2022multi} provides a unified framework for extraction and fusion, MGCMVC~\cite{ju2025multigranularity} explores multi-granularity features to maintain structural integrity, and DMCAR~\cite{xiao2026dmcar} balances shared consensus and view-specific information through a decoupled mixture-of-experts architecture with context-aware routing. 
Additionally, MAGA~\cite{bian2024multi} captures inter-view correlations through attention mechanisms, SSLNMVC~\cite{yan2025self} ensures consistency between view-specific and consensus labels using KL divergence, and GFSAF~\cite{liu2026views} addresses representation degradation by leveraging granular-ball learning to extract complementary representations alongside a multi-head cross-view attention fusion.

\subsection{Comparative Experimental Results Analysis}
\begin{table*}[t]
\centering
\caption{Comparison of clustering performance (ACC, NMI, PUR) of different methods on five multi-view datasets.}
\label{tab:result_main}
\resizebox{\linewidth}{!}{
\begin{tabular}{l ccc ccc ccc ccc ccc}
\toprule
\multirow{2}{*}{\textbf{Method}} 
& \multicolumn{3}{c}{Fashion} 
& \multicolumn{3}{c}{Synthetic3d} 
& \multicolumn{3}{c}{Cifar10} 
& \multicolumn{3}{c}{MNIST-USPS} 
& \multicolumn{3}{c}{100leaves} \\
\cmidrule(lr){2-4} \cmidrule(lr){5-7} \cmidrule(lr){8-10} \cmidrule(lr){11-13} \cmidrule(lr){14-16}
& ACC & NMI & PUR & ACC & NMI & PUR & ACC & NMI & PUR & ACC & NMI & PUR & ACC & NMI & PUR \\
\midrule
SCMVC (2024)   & 99.34 & 98.28 & 99.34 & 97.00 & 87.04 & 97.00 & 99.51 & 98.58 & 99.51 & 99.72 & 99.18 & 99.72 & 59.13 & 79.99 & 62.00 \\
CSOT (2024)    & 99.14 & 97.86 & 99.14 & 96.83 & 86.60 & 96.83 & 99.18 & 97.73 & 99.18 & 99.42 & 98.34 & 99.42 & 26.19 & 58.16 & 28.31 \\
DealMVC (2023) & 82.07 & 91.44 & 82.63 & 80.67 & 56.54 & 80.67 & 96.91 & 97.12 & 96.91 & 87.06 & 92.70 & 89.16 & 21.56 & 59.73 & 21.56 \\
SCM (2024)     & 81.86 & 78.16 & 82.03 & 94.50 & 79.86 & 94.50 & 99.27 & 97.99 & 99.27 & 98.82 & 96.74 & 98.82 & 33.25 & 65.47 & 36.06 \\
DDMVC (2025)   & 77.17 & 77.59 & 77.48 & 93.47 & 82.56 & 93.47 & 98.62 & 96.35 & 98.62 & 89.30 & 90.94 & 89.30 & 25.37 & 60.35 & 26.12 \\
MFLVC (2022)   & 99.16 & 97.95 & 99.16 & 95.67 & 83.67 & 95.67 & 99.11 & 97.72 & 99.11 & 99.60 & 98.84 & 99.60 & 24.75 & 60.81 & 26.12 \\
CVCL (2023)    & 99.31 & 98.20 & 99.31 & 89.20 & 67.17 & 89.20 & 97.26 & 95.48 & 97.26 & 99.58 & 98.77 & 99.58 & 35.24 & 67.53 & 35.24 \\
MAGA (2024)    & 98.85 & 97.26 & 98.85 & 95.50 & 82.16 & 95.50 & 98.44 & 95.74 & 98.44 & 99.24 & 97.78 & 99.24 & 43.50 & 71.61 & 46.12 \\
SEM (2023)     & 99.29 & 98.19 & 99.29 & 95.83 & 82.91 & 95.83 & 99.23 & 97.89 & 99.23 & 91.74 & 88.63 & 91.74 & 61.50 & 81.57 & 65.00 \\
SSLNMVC (2025) & 99.27 & 98.18 & 99.27 & 95.83 & 84.52 & 95.83 & 99.18 & 97.71 & 99.18 & 99.24 & 97.87 & 99.24 & 26.37 & 61.32 & 27.81 \\
MGCMVC (2025)  & 99.09 & 97.78 & 99.09 & 96.83 & 86.43 & 96.83 & 99.43 & 98.34 & 99.43 & 99.60 & 98.85 & 99.60 & 33.12 & 66.61 & 34.50 \\
MCMC (2025)    & 95.03 & 96.51 & 95.03 & 94.33 & 79.82 & 94.33 & 98.94 & 96.37 & 98.94 & 93.62 & 92.78 & 95.47 & 57.49 & 75.38 & 61.54 \\
HCLAA (2026)   & 99.37 & 98.35 & 99.37 & 97.00 & 87.01 & 97.00 & 99.53 & 98.66 & 99.53 & 99.60 & 98.82 & 99.60 & 71.56 & 85.78 & 73.44 \\
GFSAF (2026)   & 99.17 & 98.00 & 99.17 & 97.54 & 89.21 & 97.54 & 99.42 & 98.32 & 99.42 & 99.56 & 98.78 & 99.56 & 58.94 & 78.84 & 61.46 \\
DMCAR (2026)   & 99.02 & 98.14 & 99.02 & 96.27 & 88.34 & 96.27 & 99.13 & 98.28 & 99.13 & 99.43 & 98.37 & 99.43 & 60.47 & 79.46 & 60.47 \\
\midrule
\textbf{ORCA (Ours)}  & \textbf{99.52} & \textbf{98.69} & \textbf{99.52} & \textbf{97.67} & \textbf{89.56} & \textbf{97.67} & \textbf{99.54} & \textbf{98.72} & \textbf{99.54} & \textbf{99.74} & \textbf{99.26} & \textbf{99.74} & \textbf{76.81} & \textbf{89.77} & \textbf{78.63} \\
\bottomrule
\end{tabular}
}
\end{table*}

To comprehensively evaluate the effectiveness and generalization of our proposed ORCA framework, we conduct experiments on several diverse multi-view datasets. As shown in Table ~\ref{tab:result_main}, ORCA consistently achieves SOTA performance across all metrics. Specifically, on the 100leaves dataset, ORCA achieves a notable performance improvement with an ACC of 76.81\%, outperforming the second-best baseline HCLAA (71.56\%). On large-scale visual datasets such as Fashion and Cifar10, ORCA extracts highly robust representations, exceeding 99.5\% ACC and 98.6\% NMI. These remarkable results validate the efficacy of ORCA's core design. By integrating SIRF with Orthogonal Residual Semantic Discovery, the model effectively preserves raw view-specific features while explicitly extracting complementary semantic increments. Furthermore, the CCHA framework, enhanced by consensus-anchored calibration weighting, allows ORCA to exploit hierarchical complementary information across views without blind multi-scale alignment, leading to highly accurate, robust, and semantically consistent clustering outcomes.

\subsection{ Model Analysis}
\textbf{Clustering Results Visualization Analysis:}  
\begin{figure*}[htbp]
\centering

\subfloat[Raw Data]{%
    \includegraphics[width=0.24\linewidth]{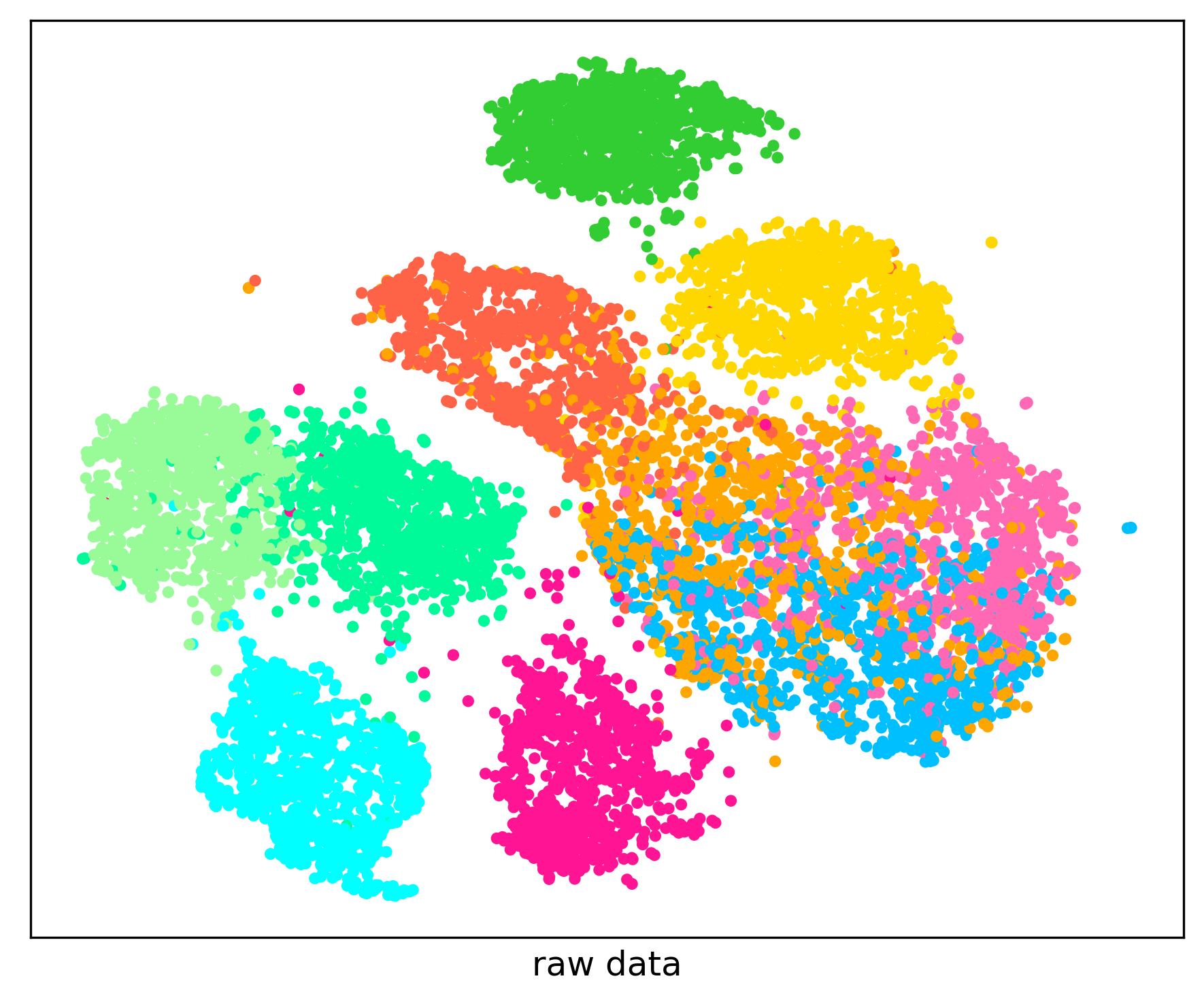}%
    \label{fig:fashion_raw}%
}
\hfill
\subfloat[Epoch 1]{%
    \includegraphics[width=0.24\linewidth]{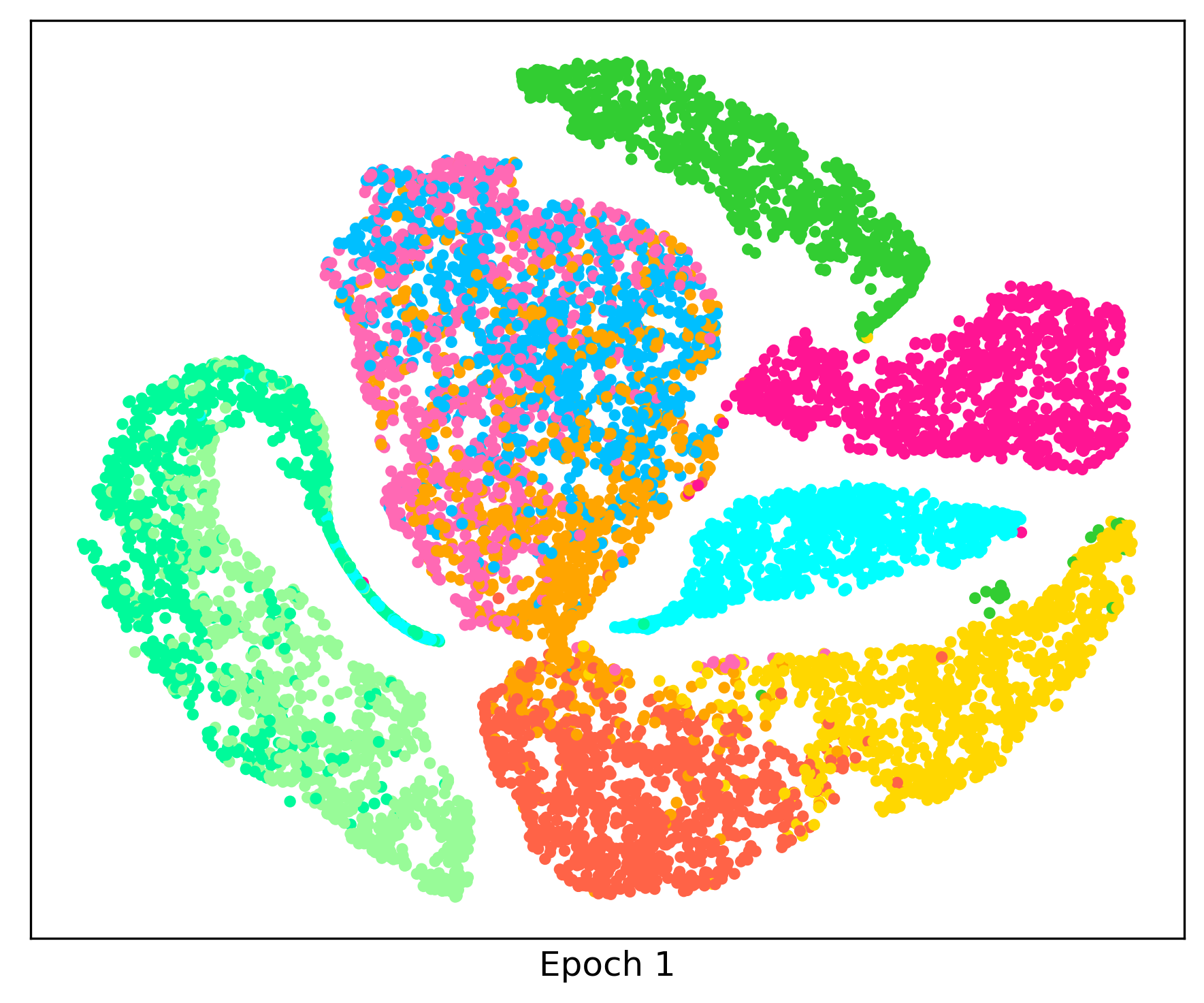}%
    \label{fig:fashion_epoch_1}%
}
\hfill
\subfloat[Epoch 50]{%
    \includegraphics[width=0.24\linewidth]{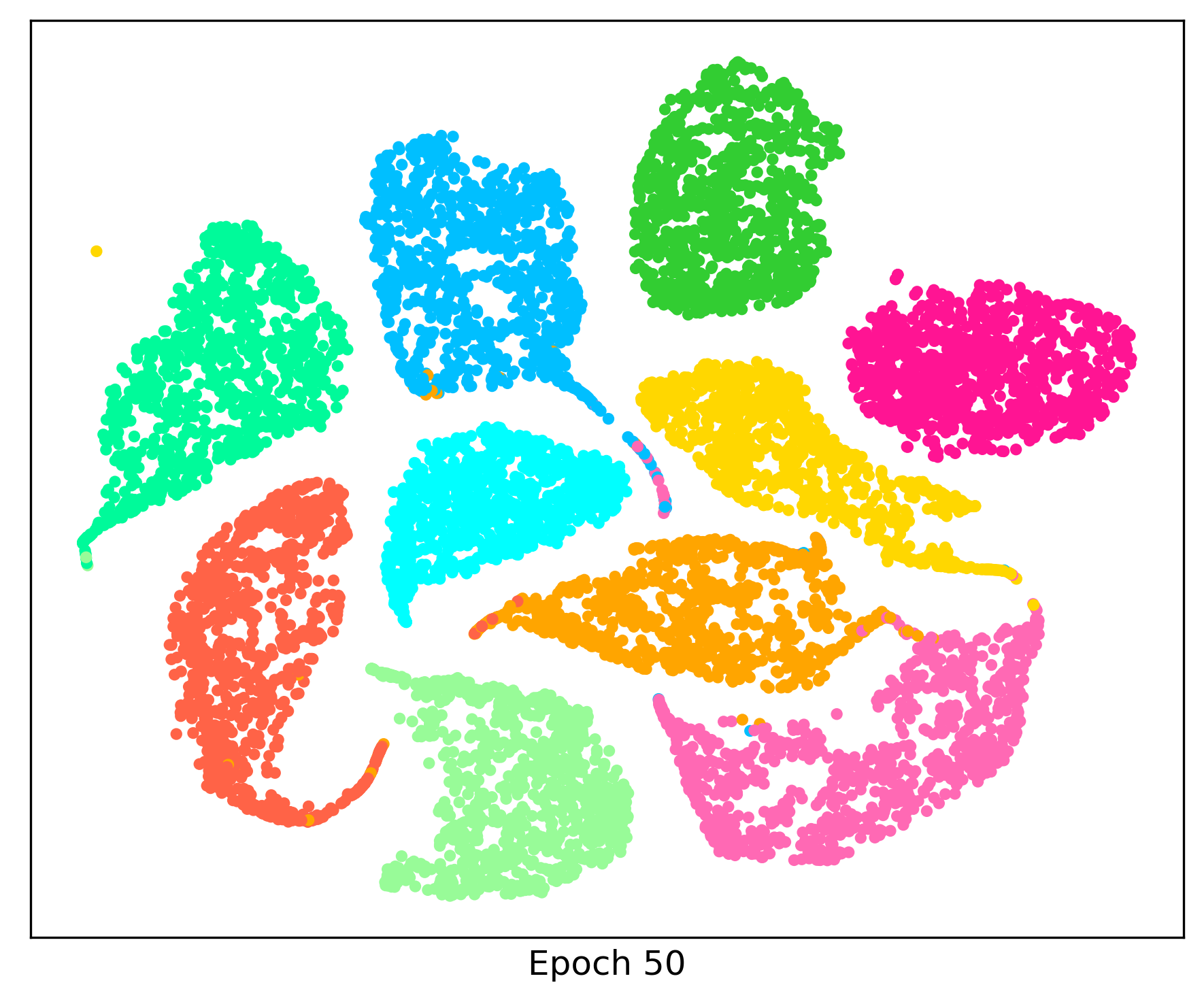}%
    \label{fig:fashion_epoch_50}%
}
\hfill
\subfloat[Epoch 100]{%
    \includegraphics[width=0.24\linewidth]{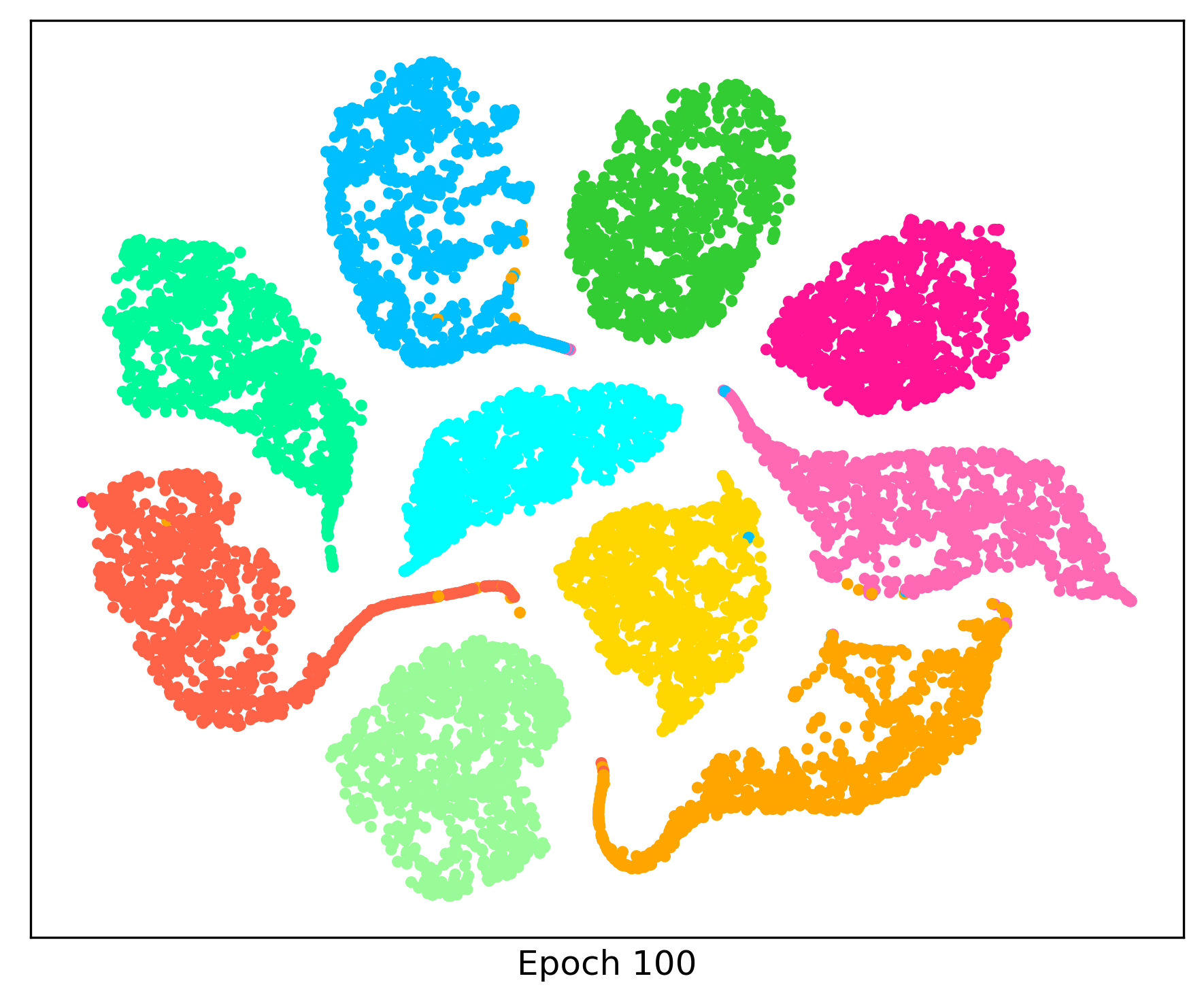}%
    \label{fig:fashion_epoch_100}%
}

\caption{Clustering visualization of the Fashion dataset at different training stages: raw data, and epochs 1, 50, and 100.}
\label{fig:tsne_fashion}
\end{figure*}
We use t-SNE to visualize the global features $\mathbf{H}$ learned by ORCA on the Fashion dataset, as shown in Figure ~\ref{fig:tsne_fashion}. The plots show that as training progresses, especially at epochs 50 and 100, clusters become more distinct with clearer class boundaries and tighter intra-class grouping. This demonstrates the effectiveness of explicitly stripping redundant shared components via semantic incremental recurrent fusion and the adaptive guidance of the consensus-calibrated hierarchical contrastive alignment mechanism in improving feature separability and structural consistency.
 
\begin{figure*}[htbp]
\centering

\subfloat[Fashion]{%
    \includegraphics[width=0.32\linewidth]{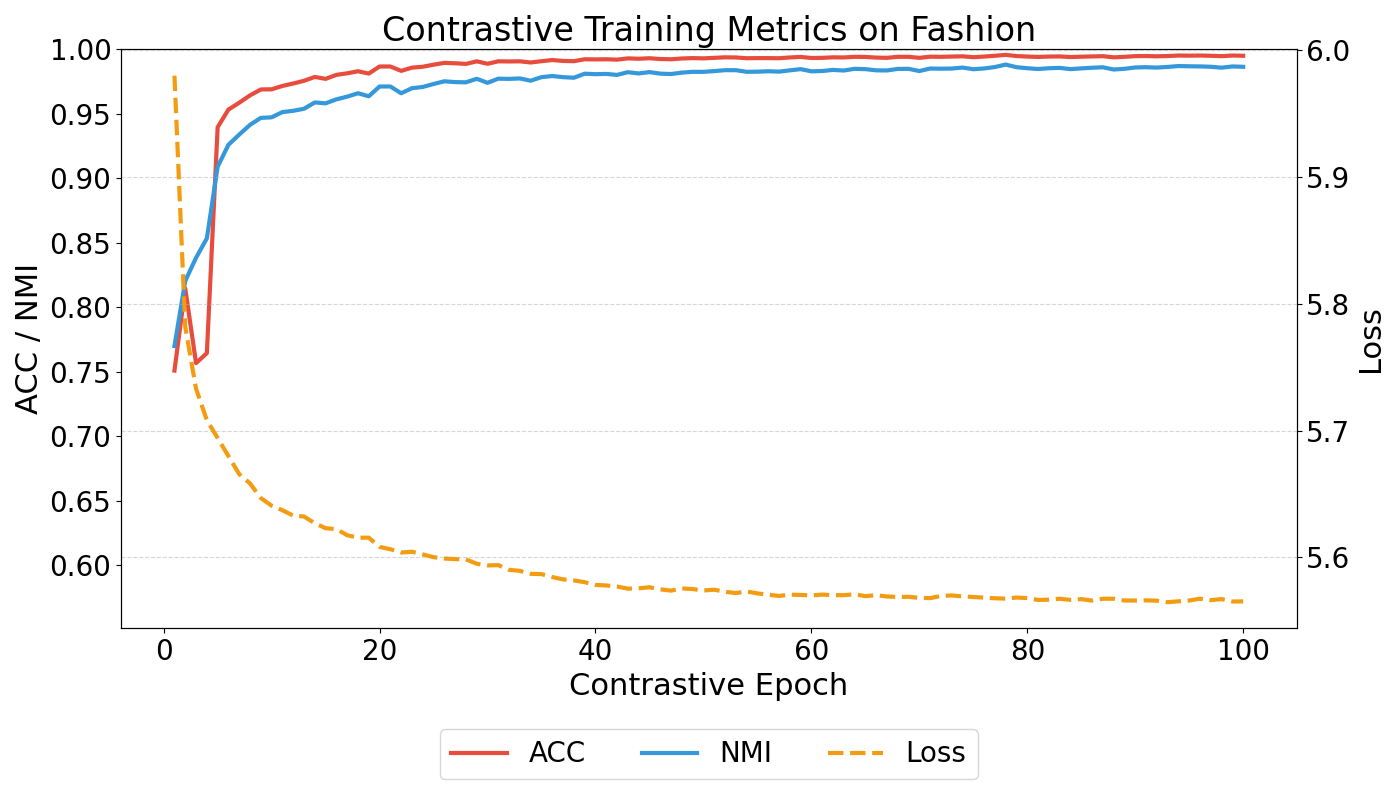}%
    \label{fig:fashion}%
}
\hfill
\subfloat[Synthetic3D]{%
    \includegraphics[width=0.32\linewidth]{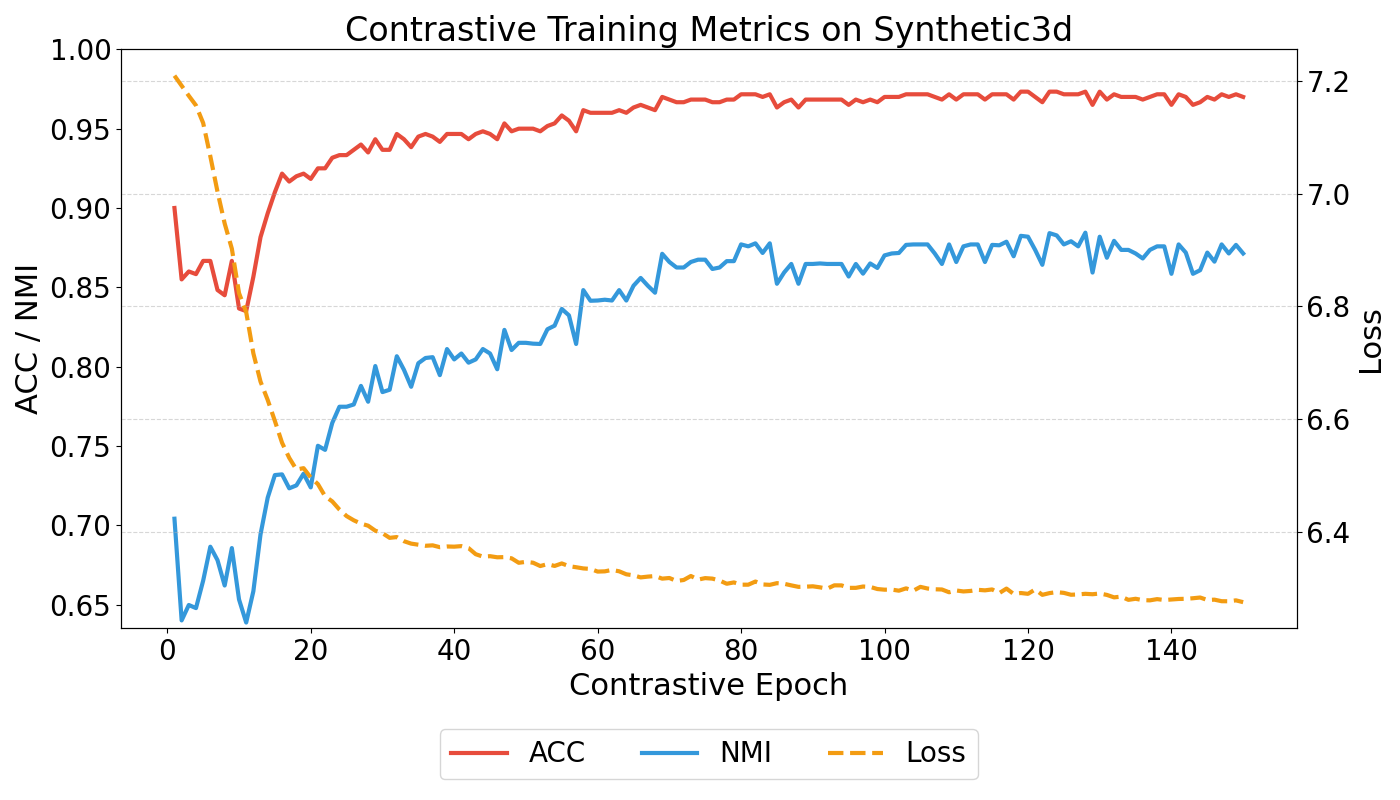}%
    \label{fig:synthetic3d}%
}
\hfill
\subfloat[MNIST-USPS]{%
    \includegraphics[width=0.32\linewidth]{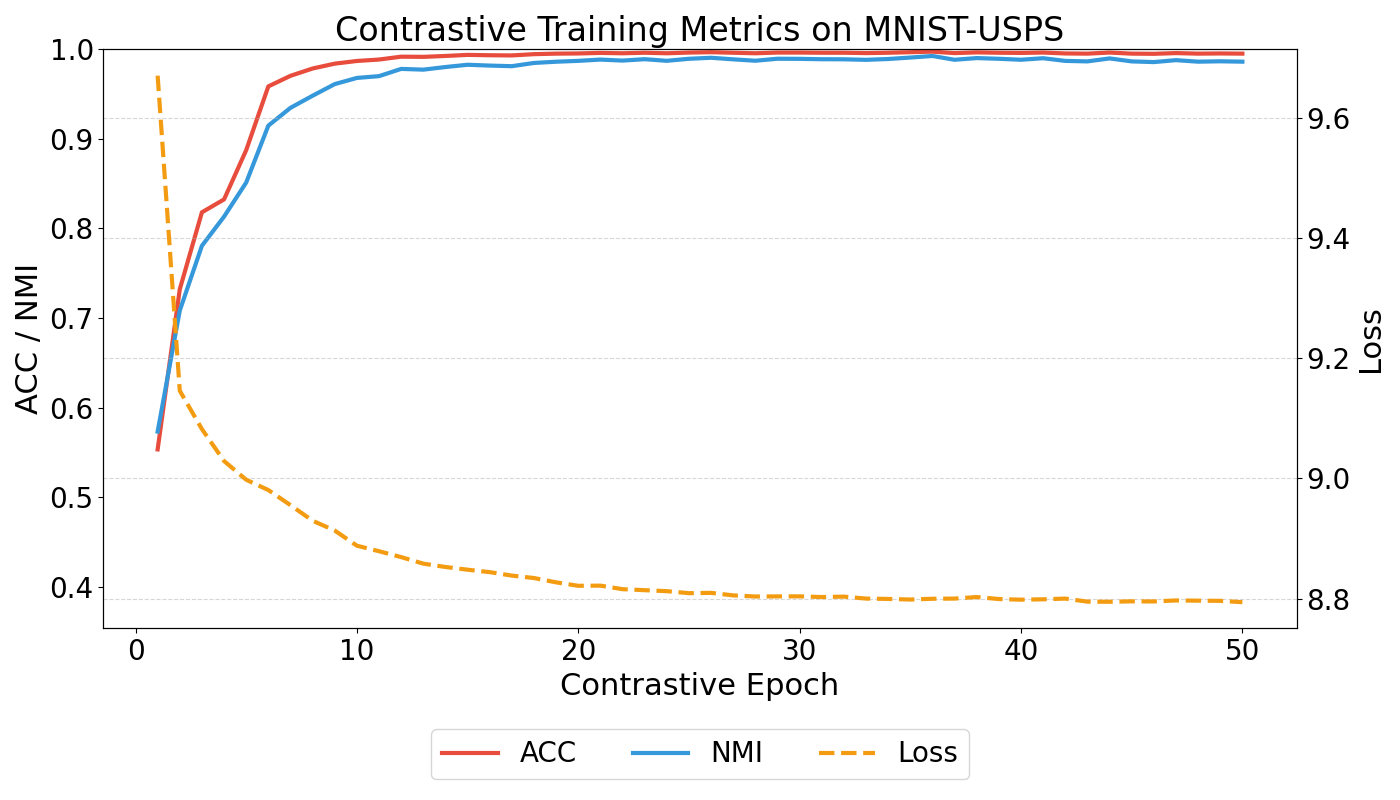}%
    \label{fig:mnist_usps}%
}

\caption{Evaluation metrics as a function of training epochs on Fashion, Synthetic3D, and MNIST-USPS datasets.}
\label{fig:convergence}
\end{figure*}

\textbf{Convergence Analysis}: 
Figure ~\ref{fig:convergence} illustrates the convergence behavior of our ORCA model on three datasets: Fashion, Synthetic3D, and MNIST-USPS. The training loss steadily decreases while clustering metrics (ACC and NMI) rapidly improve and stabilize, demonstrating effective and stable optimization. Benefiting from our proposed ORSD and CCHA mechanism, ORCA adaptively integrates multi-view features to jointly learn complementary local and global semantics. The consensus-anchored calibration weighting further facilitates hierarchical semantic interaction, enhancing adaptability and robustness on visual data. This convergence trend validates the efficiency and stability of our framework, supporting the effectiveness of our novel design.

\begin{table}[!htb]
\centering
\caption{Ablation study of different modules on five datasets.}
\label{tab:ablation_new}
\setlength{\tabcolsep}{3pt} 
\resizebox{\linewidth}{!}{
\begin{tabular}{c *{5}{c} *{15}{c}}
\toprule
\multirow{2}{*}{\textbf{Modules}}
& \multirow{2}{*}{$C_{\text{rec}}$} 
& \multirow{2}{*}{$C_H$} 
& \multirow{2}{*}{$C_{\text{ccha}}$} 
& \multirow{2}{*}{$C_W$} 
& \multirow{2}{*}{$C_{\text{orsd}}$} 
& \multicolumn{3}{c}{\textbf{Fashion}}
& \multicolumn{3}{c}{\textbf{Synthetic3D}}
& \multicolumn{3}{c}{\textbf{Cifar10}}
& \multicolumn{3}{c}{\textbf{MNIST-USPS}}
& \multicolumn{3}{c}{\textbf{100leaves}} \\
\cmidrule(lr){7-9} \cmidrule(lr){10-12} \cmidrule(lr){13-15} \cmidrule(lr){16-18} \cmidrule(lr){19-21}
& & & & &
& ACC & NMI & PUR
& ACC & NMI & PUR
& ACC & NMI & PUR
& ACC & NMI & PUR
& ACC & NMI & PUR \\
\midrule
M1 & \checkmark & & & & & 62.76 & 58.31 & 63.58 & 68.29 & 33.81 & 68.29 & 92.20 & 84.99 & 92.20 & 46.90 & 44.46 & 51.22 & 54.99 & 77.12 & 58.34 \\
M2 & \checkmark & & & & \checkmark & 96.66 & 93.74 & 81.45 & 93.88 & 64.66 & 80.88 & 96.49 & 96.66 & 93.42 & 66.33 & 57.92 & 66.33 & 62.04 & 79.27 & 63.93 \\
M3 & \checkmark & \checkmark & \checkmark & & & 98.96 & 98.26 & 98.40 & 95.89 & 85.31 & 95.89 & 98.75 & 97.59 & 98.75 & 98.83 & 97.97 & 98.83 & 63.03 & 80.35 & 64.06 \\
M4 & \checkmark & & & \checkmark & \checkmark & 95.78 & 93.06 & 80.27 & 92.79 & 59.93 & 79.88 & 96.43 & 92.89 & 85.91 & 69.33 & 60.31 & 69.33 & 63.39 & 80.06 & 64.58 \\
M5 & \checkmark & \checkmark & & & \checkmark & 99.06 & 97.99 & 99.06 & 96.59 & 87.26 & 96.59 & 99.13 & 97.97 & 99.13 & 69.92 & 61.11 & 69.92 & 61.06 & 81.70 & 65.65 \\
M6 & \checkmark & & \checkmark & & \checkmark & 96.65 & 94.80 & 84.34 & 94.91 & 69.85 & 84.84 & 96.54 & 97.15 & 94.41 & 98.00 & 96.96 & 98.00 & 62.43 & 81.57 & 64.86 \\
M7 & \checkmark & \checkmark & \checkmark & \checkmark & & 99.02 & 97.88 & 99.02 & 96.64 & 87.79 & 96.88 & 99.10 & 98.03 & 99.10 & 98.89 & 98.02 & 98.89 & 63.70 & 81.58 & 64.61 \\
M8 & \checkmark & & \checkmark & \checkmark & & 96.50 & 94.77 & 83.49 & 94.83 & 68.39 & 82.52 & 97.05 & 96.10 & 92.28 & 99.07 & 97.95 & 99.07 & 61.71 & 82.79 & 66.58 \\
M9 & \checkmark & \checkmark & & \checkmark & \checkmark & 98.21 & 96.01 & 98.21 & 95.44 & 84.02 & 95.44 & 98.89 & 97.81 & 98.89 & 98.95 & 80.01 & 98.95 & 59.23 & 78.86 & 60.50 \\
M10 & \checkmark & \checkmark & \checkmark & & \checkmark & 98.88 & 98.21 & 98.88 & 96.94 & 87.56 & 97.32 & 98.60 & 98.00 & 98.60 & 99.02 & 97.55 & 99.02 & 62.55 & 82.82 & 65.51 \\
M11 & \checkmark & & \checkmark & & & 97.13 & 94.19 & 83.38 & 94.11 & 69.92 & 84.85 & 96.81 & 96.94 & 93.24 & 97.04 & 96.97 & 97.04 & 62.72 & 82.87 & 66.16 \\
M12 & \checkmark & \checkmark & \checkmark & \checkmark & \checkmark & \textbf{99.52} & \textbf{98.69} & \textbf{99.52} & \textbf{97.67} & \textbf{89.56} & \textbf{97.67} & \textbf{99.54} & \textbf{98.72} & \textbf{99.54} & \textbf{99.74} & \textbf{99.26} & \textbf{99.74} & \textbf{76.81} & \textbf{89.77} & \textbf{78.63} \\
\bottomrule
\end{tabular}%
}
\end{table}

\subsection{Ablation Experiments}

Table ~\ref{tab:ablation_new} reports the ablation study evaluating the components of our ORCA framework. The baseline model relying solely on reconstruction loss ($C_{\text{rec}}$) exhibits limited clustering capability (e.g., 62.76\% ACC on Fashion). However, independently introducing our core modules---orthogonal residual semantic discovery ($C_{\text{orsd}}$), consensus-calibrated hierarchical contrastive alignment ($C_{\text{ccha}}$), and divergence-driven recurrent fusion ($C_H$)---yields substantial improvements. For instance, incorporating $C_{\text{orsd}}$ alone dramatically increases Fashion ACC to 96.66\%. These components demonstrate strong complementarity when combined. By further integrating the consensus-anchored calibration weighting ($C_W$) to adaptively regulate hierarchical representations, the complete ORCA framework systematically prevents topological degeneration and achieves peak performance across all datasets (Fashion: 99.52\%, Synthetic3D: 97.67\%, Cifar10: 99.54\%, MNIST-USPS: 99.74\%, 100leaves: 76.81\%). This confirms the synergistic effectiveness of our joint optimization strategy.

\section{Conclusion}
In this paper, we propose the ORCA framework to address inter-view feature redundancy and spectral degeneration in multi-view clustering. ORCA relies on two coordinated modules: SIRF, which leverages geometric projection and adaptive halting to eliminate redundancy and prevent dimensionality collapse; and CCHA, which aligns multi-scale features against a global consensus anchor to preserve highly discriminative semantics. Extensive experiments across five benchmarks demonstrate ORCA's consistent superiority over state-of-the-art methods. However, a limitation of this work is its reliance on fully aligned datasets and the lack of evaluation on large-scale real-world multi-view data, such as cross-modal industrial scenarios or web-collected data. Therefore, our future work will focus on extending ORCA to handle such complex, large-scale online clustering scenarios.

\section{Acknowledgments}

This work was supported by the National Natural Science Foundation of China (62506151).

\bibliographystyle{plainnat} 
\bibliography{lx}

\newpage 

\appendix

\section{Supplementary Experimental Details}
\label{sec:appendix_experiments}

In this supplementary section, we provide additional implementation details and experimental analyses to support the main manuscript. Specifically, we first outline the complete execution flow of the ORCA framework, detailing the systematic transition from semantic discovery to hierarchical alignment. Second, we provide qualitative visualizations of the clustering results to demonstrate the dynamic convergence of our latent representations. Finally, we conduct a parameter sensitivity analysis to evaluate the robustness of our framework under varying contrastive temperature coefficients.

\subsection{Complete Training Algorithm}

The overall training procedure of the proposed ORCA framework is summarized in Algorithm~\ref{alg:training_procedure}. The model first performs semantic incremental recurrent fusion to obtain the global semantic consensus, followed by consensus-calibrated hierarchical contrastive alignment for multi-level representation enhancement. The resulting objective is jointly optimized through end-to-end backpropagation.

\begin{algorithm}[htbp]
    \caption{Training Procedure of the Proposed ORCA Framework}
    \label{alg:training_procedure}
    \begin{algorithmic}[1]
        \Require Multi-view data $\{\mathbf{X}_v\}_{v=1}^V$, cluster number $K$, training epochs $E$, information gain threshold $\epsilon$
        \Ensure Global semantic consensus $\mathbf{H}$ and clustering assignments
        
        \For{$epoch = 1$ \textbf{to} $E$}
            \State Extract view-specific latent embeddings $\{\mathbf{z}_v\}_{v=1}^V$
            
            \State \textbf{Semantic Incremental Recurrent Fusion (SIRF):}
            \State Obtain enhanced initial representations $\mathbf{Z}^{(0)}$ via orthogonal residual semantic discovery
            \State Iteratively update $\mathbf{Z}^{(t)}$, compute information gain $\Delta\mathcal{I}^{(t)}$, and adaptively determine retention weight $\alpha_t$
            \State Aggregate the retained states to obtain global consensus $\mathbf{H}$
            
            \State \textbf{Consensus-Calibrated Hierarchical Contrastive Alignment (CCHA):}
            \State Construct the hierarchical candidate pool $\mathcal{Z}_{\mathrm{pool}}$
            \State Project candidates to representations $\mathcal{R}=\{\mathbf{R}_m\}_{m=1}^{M}$
            \State Compute consensus-guided weights and $\mathcal{L}_{\mathrm{ccha}}$ according to Eqs.~\eqref{eq:ccha_weight}--\eqref{eq:ccha_loss}
            
            \State Update all network parameters by minimizing $\mathcal{L}_{\mathrm{total}}$ in Eq.~\eqref{eq:new_total_loss}
        \EndFor
        
        \State Apply $k$-means to the optimized global consensus $\mathbf{H}$
        \State \Return $\mathbf{H}$ and the corresponding cluster labels
    \end{algorithmic}
\end{algorithm}
\subsection{Clustering Results Visualization Analysis:}

\begin{figure*}[htbp]
\centering
\subfloat[Raw Data]{%
    \includegraphics[width=0.24\linewidth]{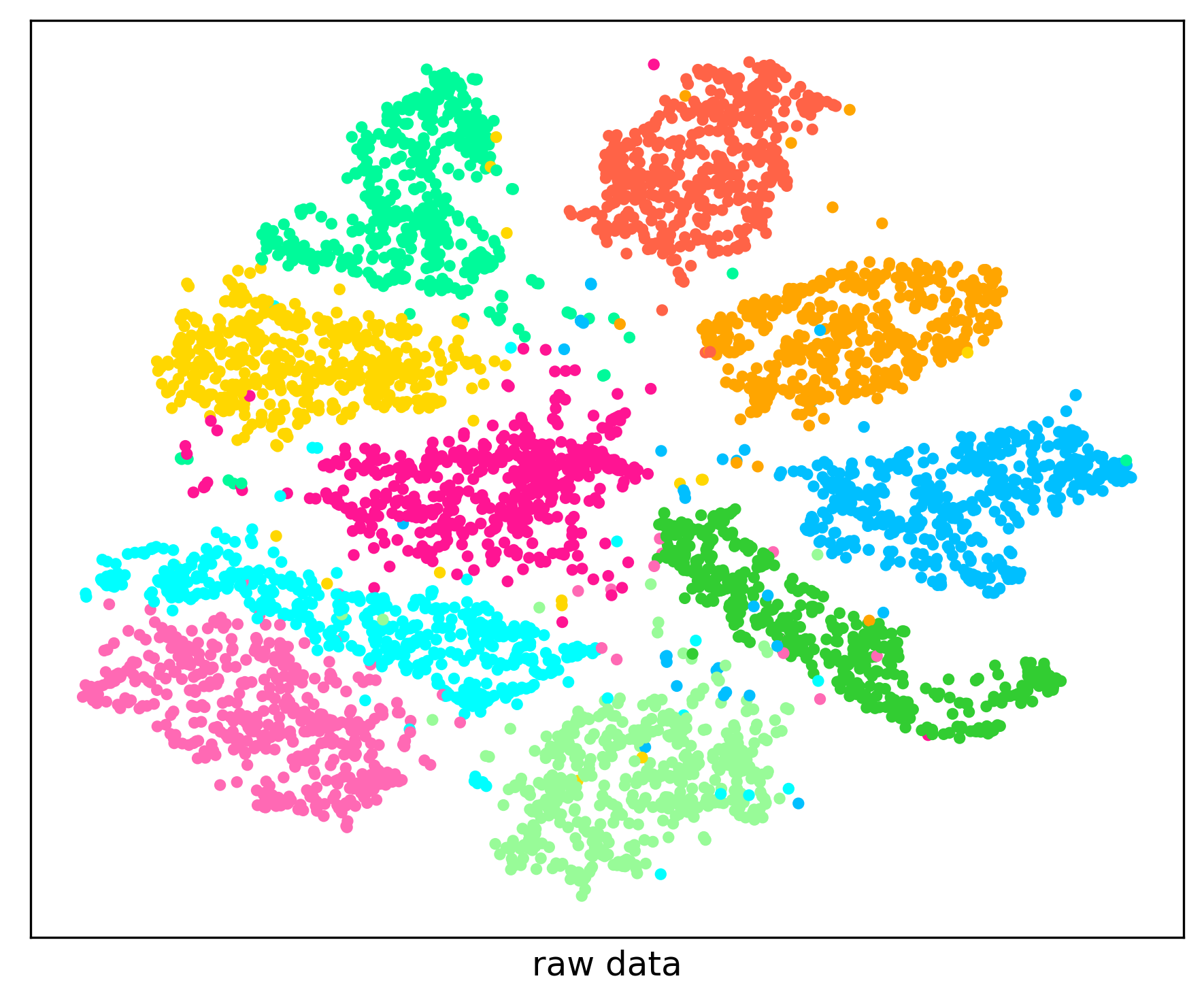}%
    \label{fig:mnist_raw}%
}
\hfill
\subfloat[Epoch 1]{%
    \includegraphics[width=0.24\linewidth]{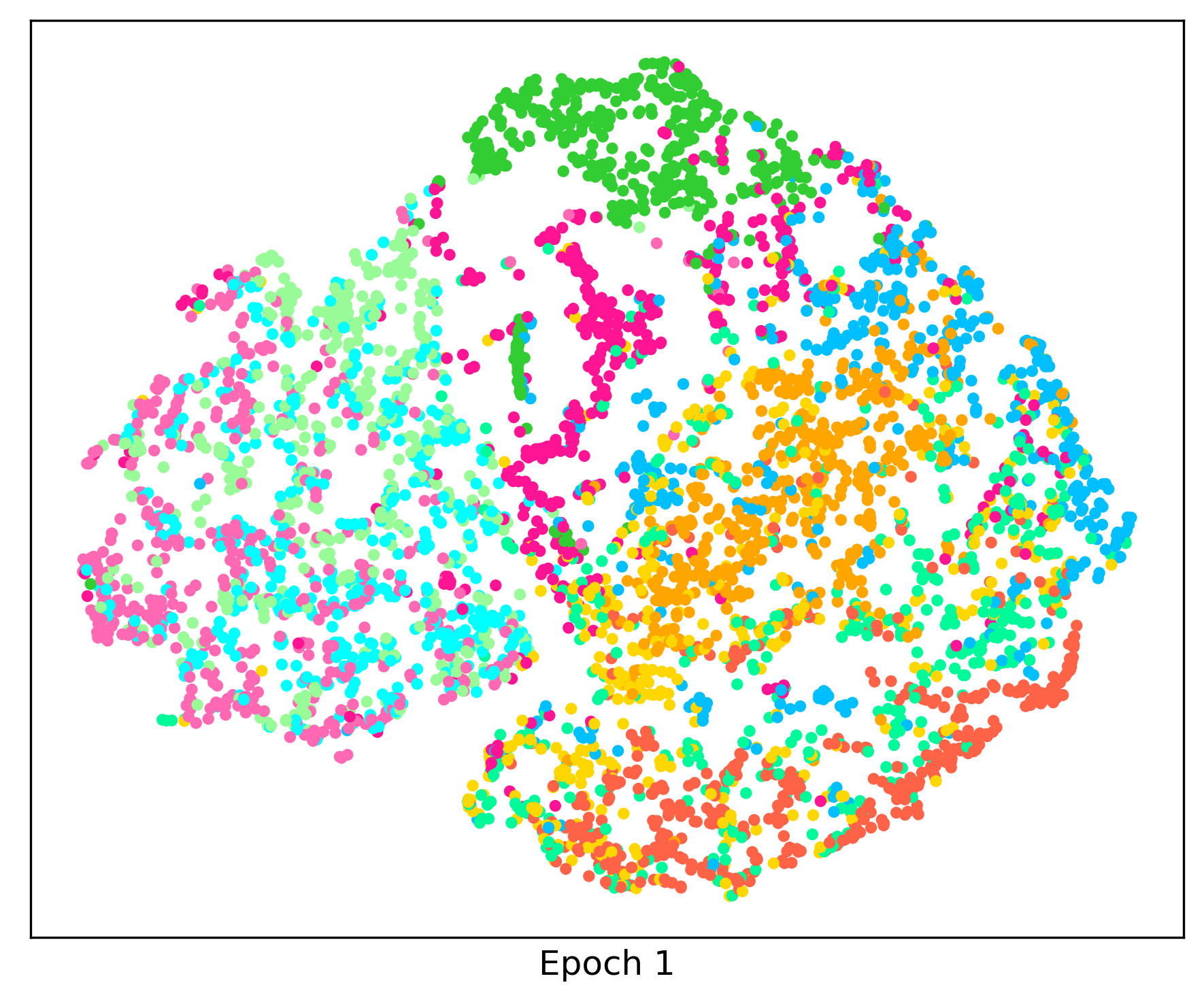}%
    \label{fig:mnist_epoch_1}%
}
\hfill
\subfloat[Epoch 20]{%
    \includegraphics[width=0.24\linewidth]{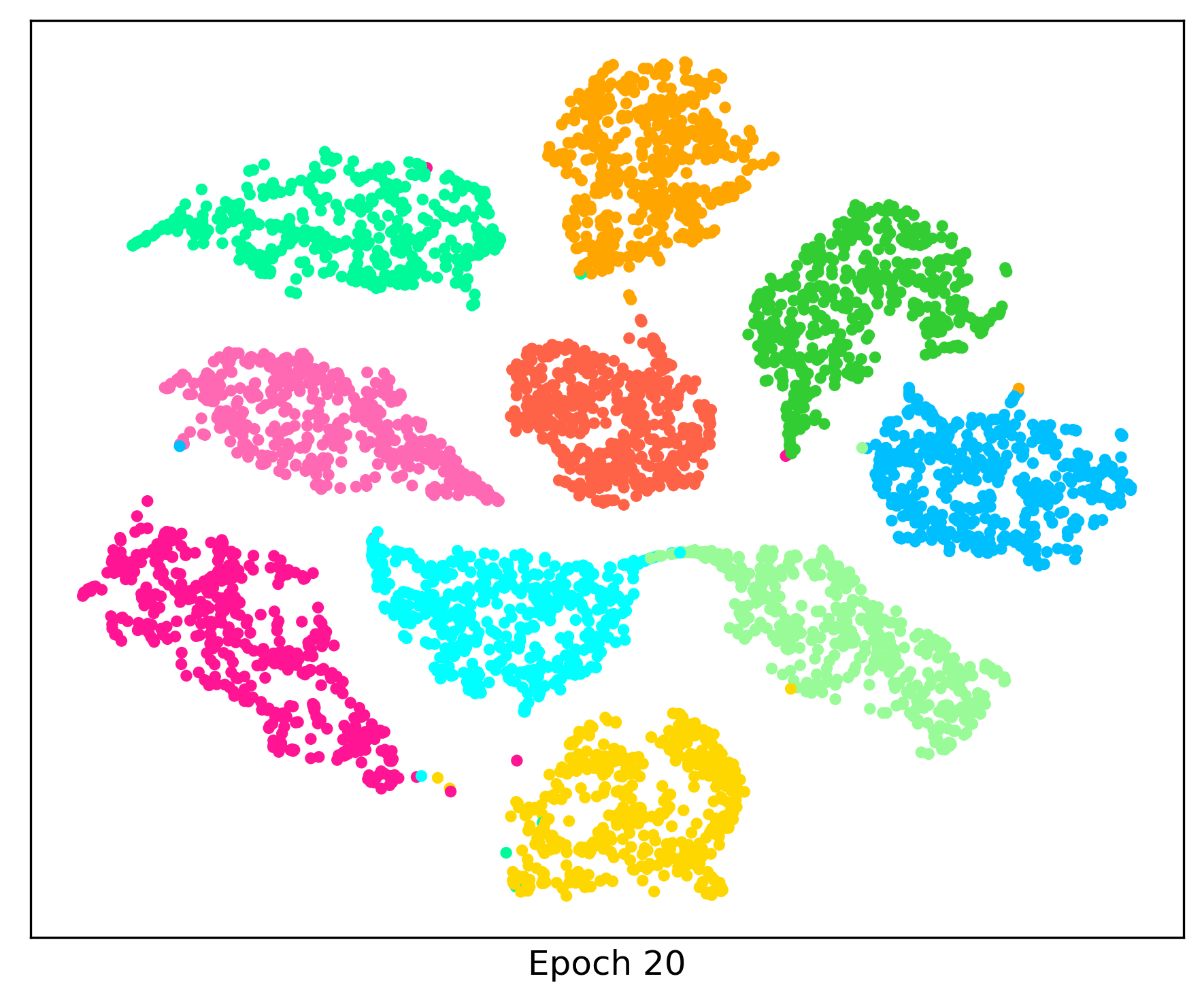}%
    \label{fig:mnist_epoch_20}%
}
\hfill
\subfloat[Epoch 50]{%
    \includegraphics[width=0.24\linewidth]{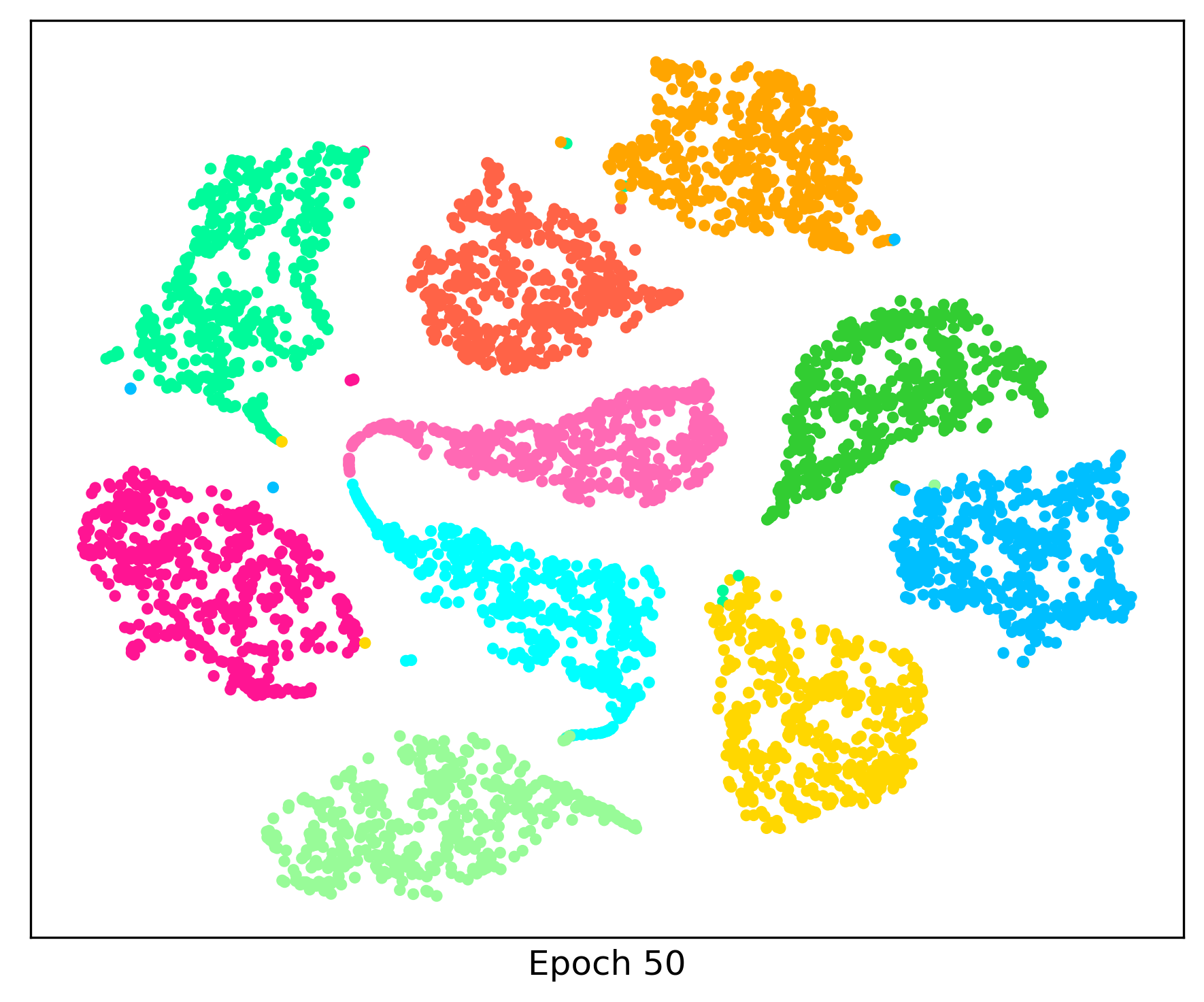}%
    \label{fig:mnist_epoch_50}%
}
\caption{Clustering visualization (t-SNE) of the MNIST-USPS dataset at different training stages: raw data, and epochs 1, 20, and 50. ORCA progressively establishes clear clustering boundaries.}
\label{fig:tsne_mnist}
\end{figure*}

To intuitively understand the representation learning process of ORCA, we map the high-dimensional latent representations into a 2D space using t-SNE. Figure ~\ref{fig:tsne_mnist} visualizes the clustering trajectory on the MNIST-USPS dataset. Initially, the raw data presents heavily overlapping and poorly separated distributions. As training progresses, ORCA progressively distills the global consensus and eliminates view-specific noise. By Epoch 50, the model yields a highly discriminative latent space characterized by compact intra-cluster structures and distinct inter-cluster margins, validating the efficacy of the proposed hierarchical alignment.

\subsection{Parameter Sensitivity Analysis}

\begin{figure*}[htbp]
    \centering
    \includegraphics[width=0.95\textwidth]{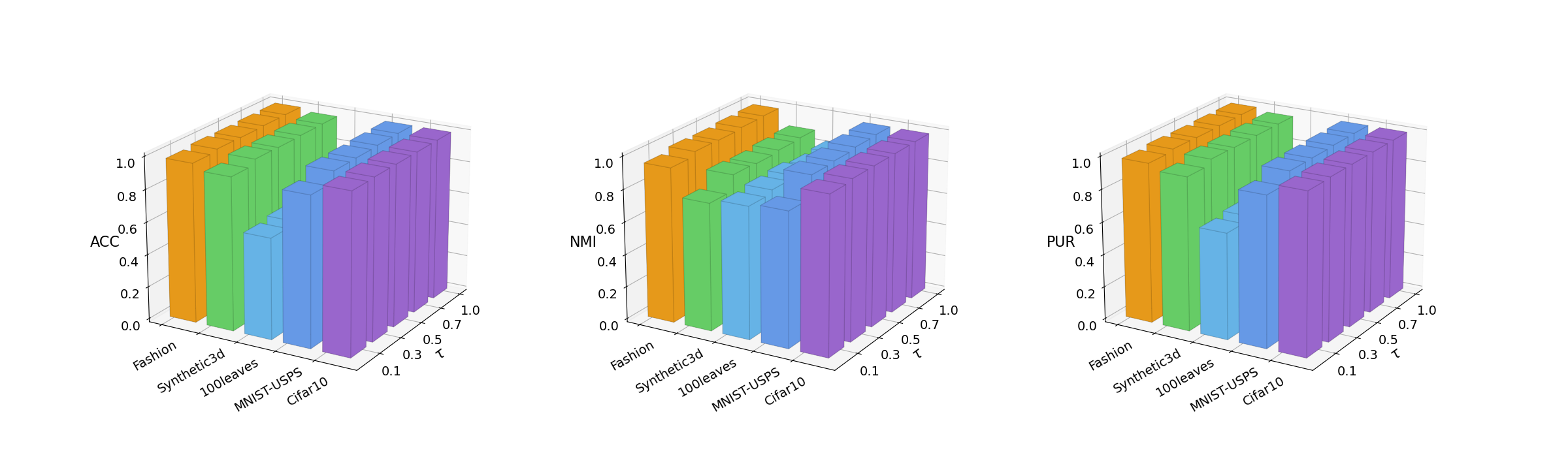}
    \caption{Analysis of clustering performance under varying temperature coefficients ($\tau$) evaluated via ACC, NMI, and PUR metrics across different datasets.}
    \label{fig:bar_metrics}
\end{figure*}

\textbf{Sensitivity to Temperature Coefficient $\tau$:} 
Figure ~\ref{fig:bar_metrics} illustrates ORCA's sensitivity to the temperature coefficient $\tau$ within the InfoNCE objective, which regulates the sharpness of the similarity distribution between the global consensus anchor $\mathbf{H}$ and the multi-scale intermediate representations $\mathbf{R}_m$. 

A very small $\tau$ (e.g., 0.1) tends to degrade clustering performance. This is because it over-compresses the representation space and excessively focuses on penalizing hard negative samples, which can disrupt global semantic consistency. Conversely, moderate values ($\tau \in [0.5, 1.0]$) consistently yield optimal results across diverse datasets by effectively balancing local feature alignment and global representation uniformity. Although different datasets (e.g., Fashion and MNIST-USPS vs. Synthetic3d and Cifar10) exhibit slightly varying sensitivities to small $\tau$ values, a well-tuned $\tau$ universally enables the consensus-anchored weighting mechanism to accurately prioritize informative hierarchical scales without metric distortion.

\section{Theoretical Analysis}

In this section, we establish a rigorous mathematical foundation for the ORCA framework, analyzing its properties through the lenses of redundancy mitigation, topological stability, and informational sufficiency. 
We first characterize the geometric properties of the orthogonal residual semantic discovery, proving that explicitly stripping the parallel projection minimizes feature redundancy and strictly bounds the condition number of the latent covariance, thereby guaranteeing an expansion of the informative subspace. 
Second, we formalize the attention-based recurrent fusion as a discrete dynamical system. We provide theoretical bounds demonstrating that without intervention, deep cross-view attention inevitably suffers from spectral degeneration. By incorporating our Adaptive Divergence-Driven Halting mechanism based on instantaneous information gain, we prove that the system rigorously avoids catastrophic dimensionality collapse, ensuring that the synthesized global consensus $\mathbf{H}$ retains maximal discriminative margin for downstream clustering.

\subsection{Information Maximization via Orthogonal Residual Projection}

A fundamental challenge in multi-view fusion is the overwhelming accumulation of redundant semantics, which distorts the latent clustering structure. We mathematically demonstrate that direct aggregation amplifies the redundant principal components, whereas our orthogonal residual injection strictly enhances the novel semantic subspace without exacerbating redundancy.

\begin{theorem}[Subspace Capacity and Redundancy Mitigation]
Let $\mathbf{\Sigma}_{\mathrm{core}} = \mathbb{E}[\mathbf{Z}_{\mathrm{core}}^{(v)} (\mathbf{Z}_{\mathrm{core}}^{(v)})^\top]$ denote the covariance matrix of the core view, and let $\mathbf{Z}_{\mathrm{other}}^{(u)}$ be a marginal view. Define the augmented representation via direct addition as $\mathbf{Z}_{\mathrm{direct}} = \mathbf{Z}_{\mathrm{core}}^{(v)} + \mathbf{Z}_{\mathrm{other}}^{(u)}$, and via ORSD's orthogonal injection as $\hat{\mathbf{Z}}^{(v)} = \mathbf{Z}_{\mathrm{core}}^{(v)} + g^{(u \to v)} \mathbf{Z}_{\bot}^{(u \to v)}$. 
The condition number of the feature covariance matrix for the ORSD formulation is strictly bounded below that of direct aggregation:
\begin{equation}
\kappa(\mathbf{\Sigma}_{\mathrm{ORSD}}) < \kappa(\mathbf{\Sigma}_{\mathrm{direct}}),
\end{equation}
improving the spectral stability of the latent space. Furthermore, the mutual information redundancy is minimized:
\begin{equation}
\mathcal{I}(\mathbf{Z}_{\mathrm{core}}^{(v)}; g^{(u \to v)} \mathbf{Z}_{\bot}^{(u \to v)}) = 0,
\end{equation}
ensuring optimal information gain.
\end{theorem}

\begin{proof}
Without loss of generality, let the latent features be zero-centered. We geometrically decompose the marginal view into its parallel and orthogonal components relative to the core view:
\begin{equation}
\mathbf{Z}_{\mathrm{other}}^{(u)} = \gamma \mathbf{Z}_{\mathrm{core}}^{(v)} + \mathbf{Z}_{\bot}^{(u \to v)}, \quad \text{where} \quad \gamma = \frac{\langle \mathbf{Z}_{\mathrm{other}}^{(u)}, \mathbf{Z}_{\mathrm{core}}^{(v)} \rangle}{\|\mathbf{Z}_{\mathrm{core}}^{(v)}\|_2^2}.
\end{equation}
Here, $\gamma$ represents the structural correlation coefficient. Under standard direct aggregation, the updated representation becomes $\mathbf{Z}_{\mathrm{direct}} = (1 + \gamma)\mathbf{Z}_{\mathrm{core}}^{(v)} + \mathbf{Z}_{\bot}^{(u \to v)}$. Given that $\mathbf{Z}_{\mathrm{core}}^{(v)}$ and $\mathbf{Z}_{\bot}^{(u \to v)}$ are strictly orthogonal ($\mathbb{E}[\mathbf{Z}_{\mathrm{core}}^{(v)} (\mathbf{Z}_{\bot}^{(u \to v)})^\top] = \mathbf{0}$), the covariance matrix of the direct aggregation expands to:
\begin{equation}
\mathbf{\Sigma}_{\mathrm{direct}} = (1+\gamma)^2 \mathbf{\Sigma}_{\mathrm{core}} + \mathbf{\Sigma}_{\bot}.
\end{equation}
The maximum eigenvalue of $\mathbf{\Sigma}_{\mathrm{direct}}$ is bounded by $\lambda_{\max}(\mathbf{\Sigma}_{\mathrm{direct}}) \geq (1+\gamma)^2 \lambda_{\max}(\mathbf{\Sigma}_{\mathrm{core}})$. This implies that the variance along the principal direction of the core view is amplified by a factor of $(1+\gamma)^2$, severely inflating the condition number $\kappa(\mathbf{\Sigma}_{\mathrm{direct}}) = \frac{\lambda_{\max}(\mathbf{\Sigma}_{\mathrm{direct}})}{\lambda_{\min}(\mathbf{\Sigma}_{\mathrm{direct}})}$ and causing redundant features to overwhelmingly dominate the latent manifold.

In contrast, ORSD explicitly isolates the orthogonal residual to compute $\hat{\mathbf{Z}}^{(v)} = \mathbf{Z}_{\mathrm{core}}^{(v)} + g^{(u \to v)} \mathbf{Z}_{\bot}^{(u \to v)}$. Its corresponding covariance matrix is strictly block-diagonalized as:
\begin{equation}
\mathbf{\Sigma}_{\mathrm{ORSD}} = \mathbf{\Sigma}_{\mathrm{core}} + (g^{(u \to v)})^2 \mathbf{\Sigma}_{\bot}.
\end{equation}
Because the variance along the core view's principal axes remains unchanged while the eigenspectrum grows only in the novel orthogonal subspace, we mathematically guarantee that $\kappa(\mathbf{\Sigma}_{\mathrm{ORSD}}) < \kappa(\mathbf{\Sigma}_{\mathrm{direct}})$, preventing spectral dominance. 

Furthermore, from an information-theoretic perspective, assuming the latent representations follow a multivariate Gaussian distribution, the mutual information is defined via the differential entropy $\mathcal{H}(\cdot)$:
\begin{equation}
\mathcal{I}(\mathbf{Z}_{\mathrm{core}}^{(v)}; \mathbf{Z}_{\bot}^{(u \to v)}) = \mathcal{H}(\mathbf{Z}_{\mathrm{core}}^{(v)}) + \mathcal{H}(\mathbf{Z}_{\bot}^{(u \to v)}) - \mathcal{H}(\mathbf{Z}_{\mathrm{core}}^{(v)}, \mathbf{Z}_{\bot}^{(u \to v)}).
\end{equation}
Since $\mathbf{Z}_{\mathrm{core}}^{(v)}$ and $\mathbf{Z}_{\bot}^{(u \to v)}$ are uncorrelated, their joint entropy equals the sum of their marginal entropies, yielding $\mathcal{I}(\mathbf{Z}_{\mathrm{core}}^{(v)}; \mathbf{Z}_{\bot}^{(u \to v)}) = 0$. This confirms that $\hat{\mathbf{Z}}^{(v)}$ spans a maximally informative subspace free of redundant overlap.
\end{proof}
\subsection{Theoretical Bounds on Spectral Degeneration via Information-Theoretic Halting}

In the context of multi-view representation learning, iterative feature fusion mechanisms inherently act as low-pass filters on the latent graph, often homogenizing discriminative cross-view semantics. To mathematically substantiate the design of the SIRF module, we first prove that unconstrained recurrent interactions inevitably drive the multi-view features toward catastrophic dimensionality collapse (i.e., over-smoothing). Subsequently, we establish that our KL-divergence guided soft-halting mechanism rigorously averts this trivial equilibrium, guaranteeing a strictly positive lower bound on the structural variance to preserve optimal clustering separability.

\begin{theorem}[Topological Stability against Dimensionality Collapse]
Let $\mathbf{Z}^{(t)} \in \mathbb{R}^{N \times d}$ be the feature matrix at iteration $t$, and let $\mathbf{L} = \mathbf{D} - \mathbf{A}$ be the graph Laplacian. The Dirichlet energy measuring latent space discriminability is defined as:
\begin{equation}
\mathcal{E}(\mathbf{Z}^{(t)}) = \frac{1}{2N} \sum_{i,j} A_{ij} \|\mathbf{z}_i^{(t)} - \mathbf{z}_j^{(t)}\|_2^2 = \frac{1}{N} \mathrm{Tr}\left( (\mathbf{Z}^{(t)})^\top \mathbf{L} \mathbf{Z}^{(t)} \right).
\end{equation}
Under unconstrained attention-based recurrent fusion, $\lim_{t \to \infty} \mathcal{E}(\mathbf{Z}^{(t)}) = 0$. 
By incorporating the dynamic soft-halting gate $\alpha_t \propto \sigma(\Delta \mathcal{I}^{(t)})$, the synthesized final representation $\mathbf{Z}_{\mathrm{final}}$ satisfies:
\begin{equation}
\mathcal{E}(\mathbf{Z}_{\mathrm{final}}) \geq \epsilon > 0,
\end{equation}
thereby successfully escaping spectral degeneration and preserving clustering boundaries.
\end{theorem}

\begin{proof}
We formulate the cross-view attention update as a Markovian graph diffusion process. Let $\mathbf{P}^{(t)}$ denote the row-stochastic attention matrix derived from the softmax operator. The core interaction is simplified as $\mathbf{Z}^{(t)} = \mathbf{P}^{(t)} \mathbf{Z}^{(t-1)} \mathbf{W}_V$. 

By the Rayleigh quotient and spectral graph theory, the propagation through the row-stochastic transition matrix inherently smooths the features. The Dirichlet energy satisfies the recursive inequality:
\begin{equation}
\mathcal{E}(\mathbf{Z}^{(t)}) \leq (1 - \lambda_2(\mathbf{L}_{\mathrm{rw}}))^2 \mathcal{E}(\mathbf{Z}^{(t-1)}),
\end{equation}
where $\lambda_2(\mathbf{L}_{\mathrm{rw}}) \in (0, 1]$ is the Fiedler eigenvalue of the random walk Laplacian. Expanding this inequality over $t$ iterations yields:
\begin{equation}
\mathcal{E}(\mathbf{Z}^{(t)}) \leq (1 - \lambda_2)^{2t} \mathcal{E}(\mathbf{Z}^{(0)}).
\end{equation}
As $t \to \infty$, the features exponentially converge to a stationary distribution (an inseparable global consensus), driving $\mathcal{E}(\mathbf{Z}^{(t)}) \to 0$. This represents a catastrophic dimensionality collapse.

To counter this, our proposed framework computes the instantaneous information gain via KL-divergence:
\begin{equation}
\Delta \mathcal{I}^{(t)} = \int_{\mathbf{Z}} \mathcal{P}(\mathbf{Z}^{(t)}) \log \frac{\mathcal{P}(\mathbf{Z}^{(t)})}{\mathcal{P}(\mathbf{Z}^{(t-1)})} d\mathbf{Z}.
\end{equation}
As the system approaches the collapsed equilibrium, the distribution shift vanishes ($\Delta \mathcal{I}^{(t)} \to 0$). The adaptive gating function $\alpha_t = \sigma(\mathbf{W}_{\alpha} \Delta \mathcal{I}^{(t)} + \mathbf{b}_{\alpha})$ forces $\alpha_t \to 0$ for saturated states. 

The final robust representation is synthesized as $\mathbf{Z}_{\mathrm{final}} = \sum_{t=1}^{T_{\mathrm{max}}} \alpha_t \mathbf{Z}^{(t)}$. Due to the convexity of the Dirichlet energy (acting as a quadratic form), we can lower-bound the energy of the final state:
\begin{equation}
\mathcal{E}(\mathbf{Z}_{\mathrm{final}}) = \frac{1}{N} \mathrm{Tr}\left( \left(\sum_{t=1}^{T_{\mathrm{max}}} \alpha_t \mathbf{Z}^{(t)}\right)^\top \mathbf{L} \left(\sum_{t=1}^{T_{\mathrm{max}}} \alpha_t \mathbf{Z}^{(t)}\right) \right) \geq \sum_{t=1}^{T_{\mathrm{max}}} \alpha_t^2 \mathcal{E}(\mathbf{Z}^{(t)}).
\end{equation}
Because the retention weights $\alpha_t$ dynamically isolate and heavily weight the early, high-energy states (where $\Delta \mathcal{I}^{(t)} \gg 0$) while strictly discarding the late collapsed states ($\alpha_t \approx 0$), the accumulated energy remains strictly positive ($\sum \alpha_t^2 \mathcal{E}(\mathbf{Z}^{(t)}) \geq \epsilon$). Thus, SIRF autonomously halts the interaction, guaranteeing that the final consensus avoids structural collapse.
\end{proof}

\newpage
\input{checklist.tex}
\end{document}

%% file: checklist.tex
\section*{NeurIPS Paper Checklist}

\begin{enumerate}

\item {\bf Claims}
    \item[] Question: Do the main claims made in the abstract and introduction accurately reflect the paper's contributions and scope?
    \item[] Answer: \answerYes{} 
    \item[] Justification: We clearly list our main contributions at the end of Section 1 (Introduction) and carefully define the scope and assumptions of our method.

\item {\bf Limitations}
    \item[] Question: Does the paper discuss the limitations of the work performed by the authors?
    \item[] Answer: \answerYes{} 
    \item[] Justification: We explicitly discuss the limitations of our proposed ORCA framework, specifically its current restriction to fully aligned datasets and offline scenarios, in the Conclusion section.

\item {\bf Theory assumptions and proofs}
    \item[] Question: For each theoretical result, does the paper provide the full set of assumptions and a complete (and correct) proof?
    \item[] Answer:\answerYes{} 
    \item[] Justification: All theoretical assumptions are clearly stated in Section X, and the complete proofs for our theorems are provided in Appendix B.

    \item {\bf Experimental result reproducibility}
    \item[] Question: Does the paper fully disclose all the information needed to reproduce the main experimental results of the paper to the extent that it affects the main claims and/or conclusions of the paper (regardless of whether the code and data are provided or not)?
    \item[] Answer:\answerYes{} 
    \item[] Justification: The paper provides sufficient details on the algorithm, datasets, and experimental setup to support reproducibility of the main results.

\item {\bf Open access to data and code}
    \item[] Question: Does the paper provide open access to the data and code, with sufficient instructions to faithfully reproduce the main experimental results, as described in supplemental material?
    \item[] Answer: \answerYes{} 
    \item[] Justification: We will upload the code package as supplementary material, and the latest
GitHub link will be provided upon paper acceptance.

\item {\bf Experimental setting/details}
    \item[] Question: Does the paper specify all the training and test details (e.g., data splits, hyperparameters, how they were chosen, type of optimizer) necessary to understand the results?
    \item[] Answer: \answerYes{} 
    \item[] Justification: The paper provides the details of data preprocessing and hyperparameter settings in the Experiments section.

\item {\bf Experiment statistical significance}
    \item[] Question: Does the paper report error bars suitably and correctly defined or other appropriate information about the statistical significance of the experiments?
    \item[] Answer: \answerNo{} 
    \item[] Justification: Most compared baseline methods do not include statistical significance in their
experiments; therefore, we follow the same practice and do not report it in our work.

\item {\bf Experiments compute resources}
    \item[] Question: For each experiment, does the paper provide sufficient information on the computer resources (type of compute workers, memory, time of execution) needed to reproduce the experiments?
    \item[] Answer: \answerYes{} 
    \item[] Justification: The computer resources is stated in the Experiments section.
    
\item {\bf Code of ethics}
    \item[] Question: Does the research conducted in the paper conform, in every respect, with the NeurIPS Code of Ethics \url{https://neurips.cc/public/EthicsGuidelines}?
    \item[] Answer: \answerYes{} 
    \item[] Justification:This paper conforms with the NeurIPS Code of Ethics.

\item {\bf Broader impacts}
    \item[] Question: Does the paper discuss both potential positive societal impacts and negative societal impacts of the work performed?
    \item[] Answer: \answerNA{} 
    \item[] Justification: The proposed algorithm has no direct societal impact. All datasets used in this
paper are publicly available, and the algorithm is solely focused on producing clustering
assignments.
    
\item {\bf Safeguards}
    \item[] Question: Does the paper describe safeguards that have been put in place for responsible release of data or models that have a high risk for misuse (e.g., pre-trained language models, image generators, or scraped datasets)?
    \item[] Answer: \answerNA{} 
    \item[] Justification: All datasets used in this paper are publicly available, so no specific safeguards
are required.

\item {\bf Licenses for existing assets}
    \item[] Question: Are the creators or original owners of assets (e.g., code, data, models), used in the paper, properly credited and are the license and terms of use explicitly mentioned and properly respected?
    \item[] Answer: \answerYes{} 
    \item[] Justification:  Original papers and datasets are properly cited.

\item {\bf New assets}
    \item[] Question: Are new assets introduced in the paper well documented and is the documentation provided alongside the assets?
    \item[] Answer: \answerNA{} 
    \item[] Justification: No new assets are introduced in the paper.

\item {\bf Crowdsourcing and research with human subjects}
    \item[] Question: For crowdsourcing experiments and research with human subjects, does the paper include the full text of instructions given to participants and screenshots, if applicable, as well as details about compensation (if any)? 
    \item[] Answer: \answerNA{} 
    \item[] Justification: No crowdsourcing or human subjects were involved in this work.

\item {\bf Institutional review board (IRB) approvals or equivalent for research with human subjects}
    \item[] Question: Does the paper describe potential risks incurred by study participants, whether such risks were disclosed to the subjects, and whether Institutional Review Board (IRB) approvals (or an equivalent approval/review based on the requirements of your country or institution) were obtained?
    \item[] Answer: \answerNA{} 
    \item[] Justification:  The paper does not involve any research with human subjects or crowdsourcing,
and therefore does not require IRB or equivalent ethical approval.

\item {\bf Declaration of LLM usage}
    \item[] Question: Does the paper describe the usage of LLMs if it is an important, original, or non-standard component of the core methods in this research? Note that if the LLM is used only for writing, editing, or formatting purposes and does \emph{not} impact the core methodology, scientific rigor, or originality of the research, declaration is not required.
    \item[] Answer:\answerNA{} 
    \item[] Justification:The core methodology does not involve the use of large language models
(LLMs).

\end{enumerate}